%% file: greenAI.tex
\documentclass{IEEEtran}
\usepackage{microtype}
\usepackage{graphicx}
\usepackage{subfigure}
\usepackage{booktabs} 
\usepackage{xcolor}
\usepackage{comment}
\usepackage{wrapfig}
\usepackage{url}
\usepackage{xurl}

\newcommand{\fb}{Facebook}
\newenvironment{itemize*}%
  {\begin{itemize}%
    \setlength{\itemsep}{0pt}%
    \setlength{\parskip}{0pt}}%
  {\end{itemize}}
\usepackage{hyperref}

\usepackage{authblk}
\newlength{\bibitemsep}
\newlength{\bibparskip}
\let\oldthebibliography\thebibliography
\renewcommand\thebibliography[1]{%
  \oldthebibliography{#1}%
  \setlength{\parskip}{\bibitemsep}%
  \setlength{\itemsep}{\bibparskip}%
}

\title{Sustainable AI: Environmental Implications, Challenges and Opportunities
}

\author{\normalsize{Carole-Jean Wu, Ramya Raghavendra, Udit Gupta, Bilge Acun, Newsha Ardalani, Kiwan Maeng,\\ Gloria Chang, Fiona Aga Behram, James Huang, Charles Bai, Michael Gschwind, Anurag Gupta, Myle Ott,\\ Anastasia Melnikov, Salvatore Candido, David Brooks, Geeta Chauhan, Benjamin Lee, Hsien-Hsin S. Lee,\\ Bugra Akyildiz, Maximilian Balandat, Joe Spisak, Ravi Jain, Mike Rabbat, Kim Hazelwood}}

\affil{Facebook AI}

\begin{document}
\maketitle
\pagestyle{plain}


\begin{abstract}
This paper explores the environmental impact of the super-linear growth trends for AI from a holistic perspective, spanning \textit{Data}, \textit{Algorithms}, and \textit{System Hardware}. 
We characterize the carbon footprint of AI computing by examining the model development cycle across industry-scale machine learning use cases and, at the same time, considering the life cycle of system hardware.
Taking a step further, we capture the operational and manufacturing carbon footprint of AI computing and present an end-to-end analysis for \textit{what} and \textit{how} hardware-software design and at-scale optimization can help reduce the overall carbon footprint of AI.
Based on the industry experience and lessons learned, we share the key challenges and chart out important development directions across the many dimensions of AI. 
We hope the key messages and insights presented in this paper can inspire the community to advance the field of AI in an environmentally-responsible manner.
\end{abstract}

\input{tex/introduction}

\input{tex/model-life-cycle}
\input{tex/cf-characterization}

\input{tex/futureofai}

\input{tex/takeaways}
\input{tex/conclusion}

\section*{Acknowledgement}

We would like to thank 
Nikhil Gupta,
Lei Tian, 
Weiyi Zheng, 
Manisha Jain,
Adnan Aziz,
and Adam Lerer for their feedback on many iterations of this draft, and in-depth technical discussions around building efficient infrastructure and platforms;  
Adina Williams,
Emily Dinan,
Mona Diab,
Ashkan Yousefpour for the valuable discussions and insights on AI and environmental responsibility; 
Mark Zhou,
Niket Agarwal,
Jongsoo Park,
Michael Anderson,
Xiaodong Wang; 
Yatharth Saraf,
Hagay Lupesco, Jigar Desai, Joelle Pineau, 
Ram Valliyappan, Rajesh Mosur,
Ananth Sankarnarayanan and
Eytan Bakshy for their leadership and vision without which this work would not have been possible.




\bibliographystyle{ieeetr}
\bibliography{greenai}

\newpage
\appendix

\section{An Sustainability Mindset for AI}
\label{sec:appendix-efficiiency-mindset}

Despite the recent calls-to-action~\cite{Strubell:arxiv:2019,Lacoste:arxiv:2019,Henderson:arxiv:2020,Bender:facct:2021}, the overall community remains under-invested in research that aims at deeply understanding and minimizing the cost of AI.
There are several factors that may have contributed to the current state of AI:

\begin{itemize}

\setlength\itemsep{0em}
    \item {\bf Lack of incentives:} Over 90\% of the ML publications only focus on model accuracy improvements at the expense of efficiency~\cite{Schwartz:arxiv:2019}. Challenges\footnote{Efficient Open-Domain Question Answering (\url{https://efficientqa.github.io/}), SustaiNLP: Simple and Efficient Natural Language Processing  (\url{https://sites.google.com/view/sustainlp2020/home}), and WMT: Machine Translation Efficiency Task (\url{http://www.statmt.org/wmt21/efficiency-task.html}).} incentivize investment into efficient approaches.
    \item {\bf Lack of common tools:} There is no standard telemetry in place to provide accurate, reliable energy and carbon footprint measurement. The measurement methodology is complex --- factors, such as datacenter infrastructures, hardware architectures, energy sources, can perturb the final measure easily.
    \item {\bf Lack of normalization factors:} Algorithmic progress in ML is often presented in some measure of model accuracy, e.g., BLEU, points, ELO, cross-entropy loss, but without considering resource requirement as a normalization factor, e.g., the number of \\CPU/GPU/TPU hours used, the overall energy consumption and/or carbon footprint required.   
    \item {\bf Platform fragmentation:} Implementation details can have a significant impact on real-world efficiency, but best practices remain elusive and platform fragmentation prevents performance and efficiency portability across model development.

\end{itemize}

\section{Additional Opportunities for AI Research and Development}
\label{sec:additional-opportunities}
\vspace{0.4cm}

\begin{figure}[t]
    \centering
    \includegraphics[width=\linewidth]{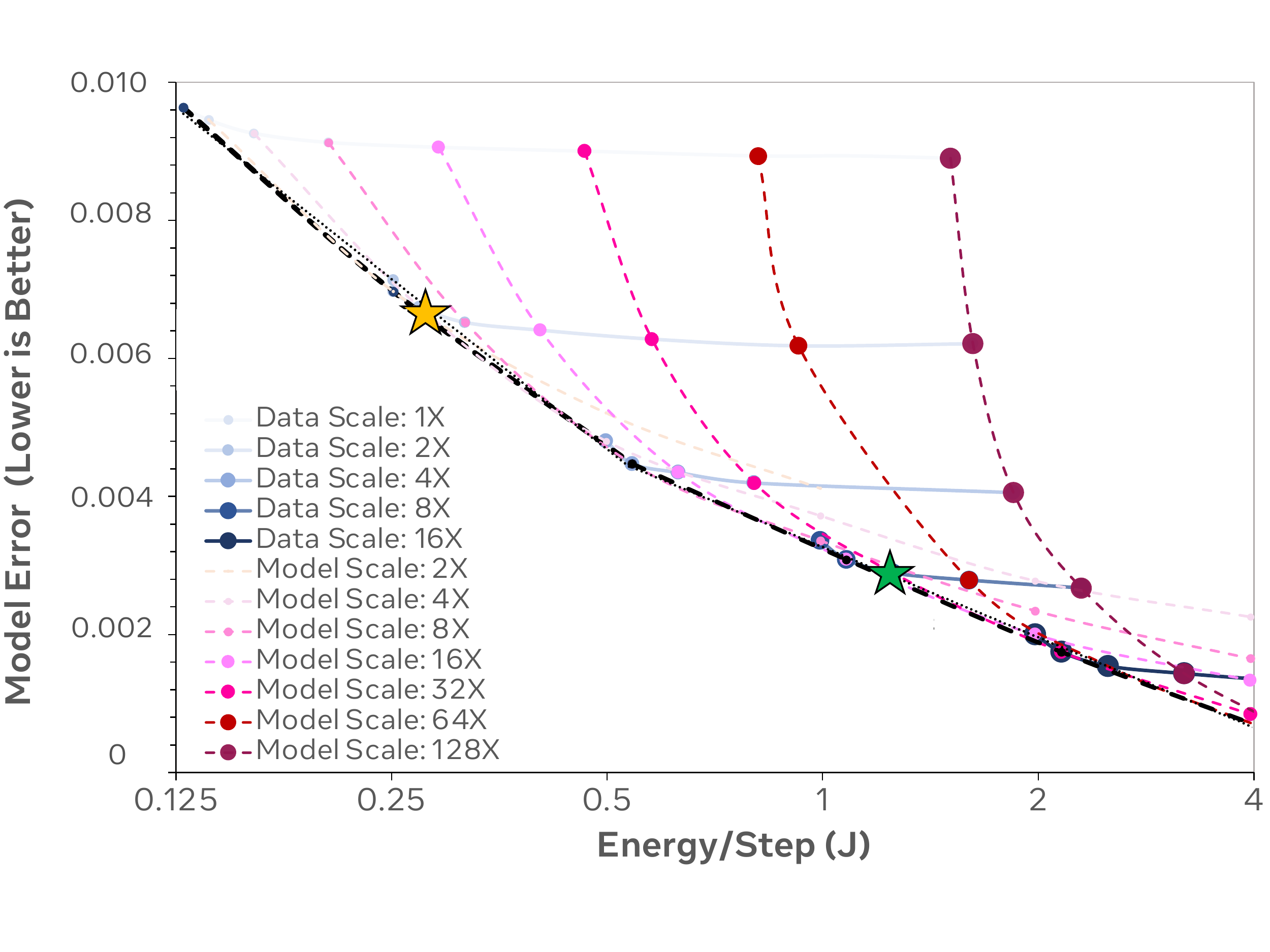}
    \vspace{-1cm}
    \caption{Model quality of recommendation use cases improves as we scale up the amount of data and/or the number of model parameters (e.g., embedding cardinality or dimension), leading to higher energy and carbon footprint. Maximizing model accuracy for the specific recommendation use case comes with significant energy cost --- Roughly 4$\times$ energy saving can be achieved with only 0.004 model quality degradation (green vs. yellow stars).}
    \label{fig:energy-scaling}
    \vspace{-0.4cm}
\end{figure}

\subsection{Data Utilization Efficiency}
\label{sec:appendix-data-efficiency}

Figure~\ref{fig:energy-scaling} depicts energy footprint reduction potential when data and model scaling is performed in tandem. The x-axis represents the energy footprint required per training step whereas the y-axis represents model error. 
The \textbf{blue} solid lines capture model size scaling (through embedding hash scaling) while the training data set size is kept fixed. 
Each line corresponds to a different data set size, in an increasing order from top to bottom.
The points within each line represent different model (embedding) sizes, in an increasing order from left to right. 
The \textbf{red} dashed lines capture data scaling while the model size is kept fixed. 
Each line corresponds to a different embedding hash size, in an increasing order from left to right.
The points within each line represent different data sizes, in an increasing order from top to bottom. 
The dashed black line captures the performance scaling trend as we scale data and model sizes in tandem. This represents the energy-optimal scaling approach.

Scaling data sizes or model sizes independently deviates from the energy-optimal trend.
We highlight two energy-optimal settings along the Pareto-frontier curve.
The yellow star uses the scaling setting of \textit{Data scaling 2$\times$} and \textit{Model scaling 2$\times$} whereas the green star adopts the setting of \textit{Data scaling 8$\times$} and \textit{Model scaling 16$\times$}.
The yellow star consumes roughly 4$\times$ lower energy as compared to the green star with only 0.004 model quality degradation in Normalized Entropy. 
Overall model quality performance has a (diminishing) power-law relationship with the corresponding energy consumption and the power of the power law is extremely small (0.002-0.004). This means achieving higher model quality through model-data scaling for recommendation use cases incurs significant energy cost.


\subsection{Efficient, Environmentally-Sustainable AI Systems}
\label{sec:appendix-system-efficiency}

\textbf{Disaggregating Machine Learning Pipeline Stages:} As depicted in Figure~\ref{fig:ml_lifecycle}, the overall training throughput efficiency for large-scale ML models depends on the throughput performance of both \textit{data ingestion and pre-processing} and \textit{model training}. Disaggregating the data ingestion and pre-processing stage of the machine learning pipeline from model training is the de-facto approach for industry-scale machine learning model training. This allows training accelerator, network and storage I/O bandwidth utilization to scale independently, thereby increasing the overall model training throughput by 56\%~\cite{Zhao:arxiv:2021}. Disaggregation with well-designed check-pointing support~\cite{Maeng:arxiv:2021,Eisenman:arxiv:2021} improves training fault tolerance as well. By doing so, failure on nodes that are responsible for data ingestion and pre-processing can be recovered efficiently without requiring re-runs of the entire training experiment. From a sustainability perspective, disaggregating the data storage and ingestion stage from model training maximizes infrastructure efficiency by \textit{using less system resources to achieve higher training throughput}, resulting in lower embodied carbon footprint. By increasing fault tolerance, the operational carbon footprint is reduced at the same time.  

\textbf{Fault-Tolerant AI Systems and Hardware:}
One way to amortize the rising embodied carbon cost of AI infrastructures is to extend hardware lifetime. However, hardware ages --- depending on the wear-out characteristics, increasingly more errors can surface over time and result in \textit{silent data corruption}, leading to erroneous computation, model accuracy degradation, non-deterministic ML execution, or fatal system failure. In a large fleet of processors, silent data corruption can occur frequently enough to have disruptive impact on service productivity~\cite{Dixit:arxiv:2021,Hochschild:hotos:2021}. Decommissioning an AI system entirely because of hardware faults is expensive from the perspective of resource and environmental footprints. System architects can design differential reliability levels for micro architectural components on an AI system depending on the ML model execution characteristics. Alternatively, algorithmic fault tolerance can be built into deep learning programming frameworks to provide a code execution path that is cognizant of hardware wear-out characteristics.

\textbf{On-Device Learning:} 
Federated learning and optimization can result in a non-negligible amount of carbon emissions at the edge, similar to the carbon footprint of training $Transformer_{Big}$~\cite{Patterson:arxiv:2021}.
Figure~\ref{fig:fl_carbon} shows that the federated learning and optimization process emits non-negligible carbon at the edge due to both computation and wireless communication during the process. 
To estimate the carbon emission, we used a similar methodology to~\cite{flcarbon}. We collected the 90-day log data for federated learning production use cases at \fb, which recorded the time spent on computation, data downloading, and data uploading per client device. We multiplied the computation time with the estimated device power and upload/download time with the estimated router power, and omitted other energy. We assumed a device power of 3W and a router power of 7.5W~\cite{phone_ml_energy, flcarbon}.
Model training on client edge devices is inherently less energy-efficient because of the high wireless communication overheads, sub-optimal training data distribution in individual client devices~\cite{flcarbon}, large degree of system heterogeneity among client edge devices, and highly-fragmented edge device architectures that make system-level optimization significantly more challenging~\cite{wu:hpca:2019}. Note, the wireless communication energy cost takes up a significant portion of the overall energy footprint of federated learning, making energy footprint optimization on communication important.

\subsection{Efficiency and Self-Supervised Learning}
\label{sec:ssl}
\input{tex/ssl}

\end{document}

%% file: tex/introduction.tex
\section{Introduction}
\label{sec:introduction}

\begin{figure}[t]
    \centering
    \includegraphics[width=\linewidth]{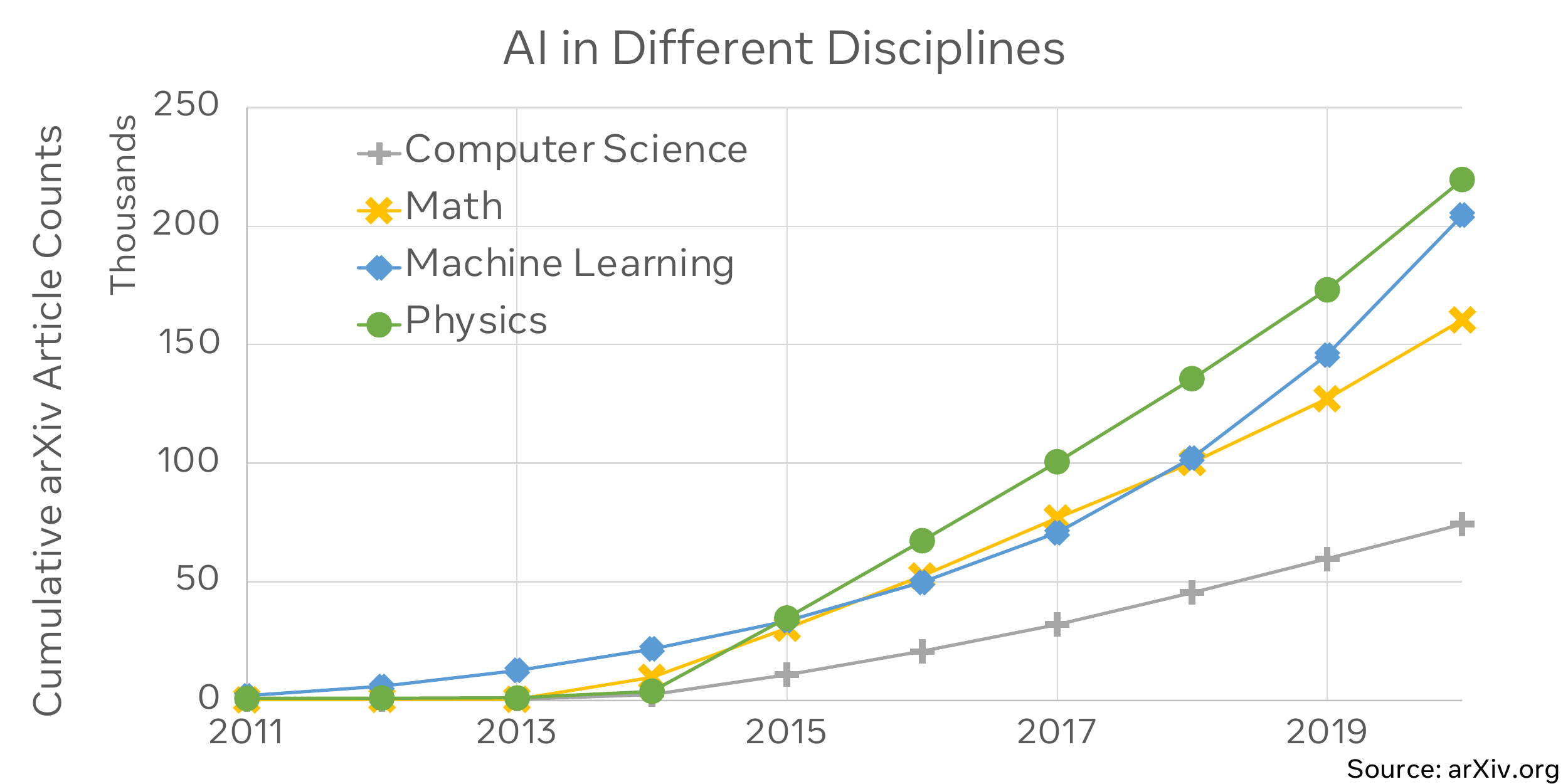}
    \caption{The growth of ML is exceeding that of many other scientific disciplines. Significant research growth in machine learning is observed in recent years as illustrated by the increasing cumulative number of papers published in machine learning with respect to other scientific disciplines based on the monthly count
    (y-axis measures the cumulative number of articles on arXiv).
    }
    \vspace{-0.5cm}
    \label{fig:ai-growth}
\end{figure}

Artificial Intelligence (AI) is one of the fastest growing domains spanning research and product development and significant investment in AI is taking place across nearly every industry, policy, and academic research. This investment in AI has also stimulated novel applications in domains such as science, medicine, finance, and education. Figure~\ref{fig:ai-growth} analyzes the number of papers published within the scientific disciplines, illustrating the growth trend in recent years\footnote{Based on monthly counts, Figure~\ref{fig:ai-growth} estimates the cumulative number of papers published per category on the arXiv database.}.

AI plays an instrumental role to push the boundaries of knowledge and sparks novel, more efficient approaches to conventional tasks.
AI is applied to predict protein structures radically better than previous methods. It has the potential to revolutionize biological sciences by providing in-silico methods for tasks only possible in a physical laboratory setting~\cite{AlphaFold}. 
AI is demonstrated to achieve human-level conversation tasks, such as the Blender Bot~\cite{Komeili:arxiv:2021}, and play games at superhuman levels, such as AlphaZero \cite{AlphaZero}.
AI is used to discover new electrocatalysts for efficient and scalable ways to store and utilize renewable energy~\cite{open-catalyst}, predicting renewable energy availability in advance to improve energy utilization~\cite{AI-load-shaping}, 
operating hyperscale data centers efficiently~\cite{google-cloud}, 
growing plants using less natural resources~\cite{robot-farms}, and, at the same time,
being used to tackle climate changes~\cite{rolnick:arxiv:2019,Nishant:IJIM:2020}.
It is projected that, in the next five years, the market for AI will increase by 10$\times$ into hundreds of billions of dollars~\cite{AI-market}. 
All of these investments in research, development, and deployment have led to a super-linear growth in AI data, models, and infrastructure capacity.
With the dramatic growth of AI, it is imperative to understand the environmental implications, challenges, and opportunities of this nascent technology. This is because technologies tend to create a self-accelerating growth cycle, putting new demands on the environment. 

This work explores the environmental impact of AI from a \textit{holistic} perspective. 
More specifically, we present the challenges and opportunities to designing sustainable AI computing across the key phases of the machine learning (ML) development process --- \textit{Data}, \textit{Experimentation}, \textit{Training}, and \textit{Inference} --- for a variety of AI use cases at \fb, such as vision, language, speech, recommendation and ranking. The solution space spans across our fleet of datacenters and on-device computing. 
Given particular use cases, we consider the impact of AI \textit{data}, \textit{algorithms}, and \textit{system hardware}.
Finally, we consider emissions across the life cycle of hardware systems, from manufacturing to operational use. 

\begin{figure*}[t]
    \centering
    \includegraphics[width=\linewidth]{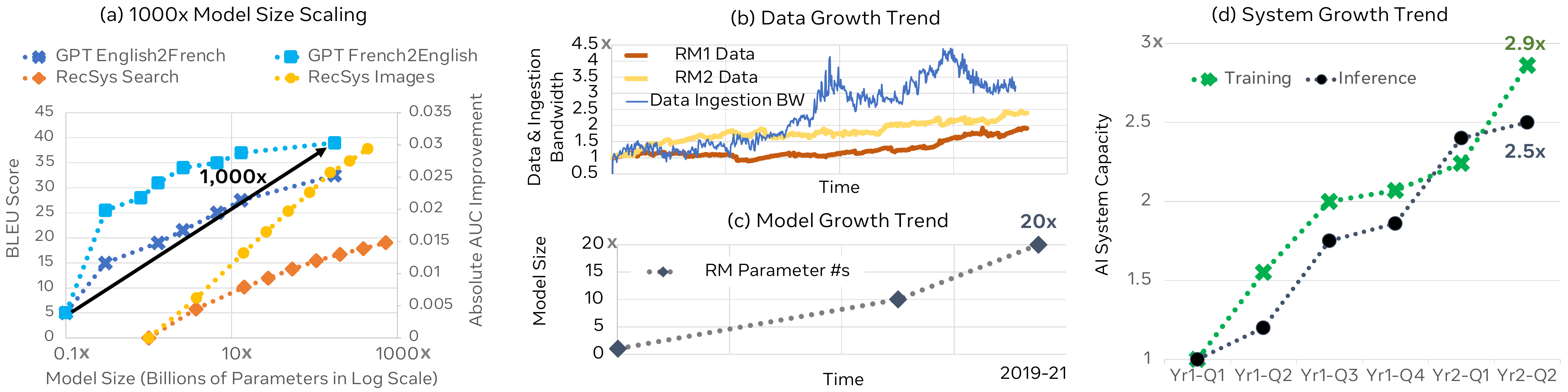}
    \caption{Deep learning has witnessed an exponential growth in data, model parameters, and system resources over the recent years.
    (a) The $1000\times$ model size growth has led to higher model accuracy for various ML tasks. For example, with GPT-3, to increase the model quality BLEU score from 5 to 40 requires a model $1,000\times$ larger in size. 
    (b) At \fb, the amount of data for recommendation use cases has roughly doubled between 2019 and 2021, leading to 3.2 times increase in the data ingestion bandwidth demand.
    (c) 
    \fb's recommendation and ranking model sizes have increased by 20 times during the same time period~\cite{Mudigere:scaling-training:2021}.
    (d) The explosive growth in AI has driven $2.9\times$ and $2.5\times$ capacity increases for AI training and inference, respectively.
    }
    \vspace{-0.25cm}
    \label{fig:data-model-system-growth}
\end{figure*}

\textbf{AI Data Growth.}
In the past decade, we have seen an exponential increase in AI training data and model capacity.
Figure~\ref{fig:data-model-system-growth}(b) illustrates that the amount of training data at Facebook for two recommendation use cases --- one of the fastest growing areas of ML usage at \fb --- has increased by 2.4$\times$ and 1.9$\times$ in the last two years, reaching exabyte scale.
The increase in data size has led to a 3.2$\times$ increase in data ingestion bandwidth demand.
Given this increase, data storage and the ingestion pipeline accounts for a significant portion of the infrastructure and power capacity compared to ML training and end-to-end machine learning life cycles.


\textbf{AI Model Growth.}
The ever-increasing data volume has also driven a super-linear trend in model size growth. 
Figure~\ref{fig:data-model-system-growth}(a) depicts the $1000\times$ model size increase for GPT3-based language translation tasks~\cite{Hernandez:arxiv:2020,brown:arxiv:2020}, whereas for Baidu's search engine, the model of $1000\times$ larger in size improves accuracy in AUC by 0.030. Despite small, the accuracy improvement can lead to significantly higher-quality search outcomes~\cite{g-search}. 
Similarly, Figure~\ref{fig:data-model-system-growth}(c) illustrates that between 2019 and 2021, the size of recommendation models at Facebook has increased by 20$\times$~\cite{Yi2018FactorizedDR,zhao2020distributed,lui2020understanding, Mudigere:scaling-training:2021}.
Despite the large increase in model sizes, the memory capacity of GPU-based AI accelerators, e.g. 32GB (NVIDIA V100, 2018) to 80GB (NVIDIA A100, 2021), has increased by $<2\times$ every 2 years.
The resource requirements for strong AI scaling clearly outpaces that of system hardware.

\textbf{AI Infrastructure Growth.}
The strong performance scaling demand for ML motivates a variety of \textit{scale-out} solutions~\cite{Mudigere:scaling-training:2021,Rajbhandari:zero:2021} by leveraging parallelism at scale with a massive collection of training accelerators.
Figure~\ref{fig:data-model-system-growth}(d) illustrates that the explosive growth in AI use cases at Facebook has driven $2.9\times$ increase in AI training infrastructure capacity over the 1.5 years.
In addition, we observe trillions of inference per day across \fb's data centers---more than doubling in the past 3 years. 
The increase in inference demands has also led to an $2.5\times$ increase in AI inference infrastructure capacity. 
Last but not least, the carbon footprint of AI goes beyond its \textit{operational} energy consumption. 
The \textit{embodied} carbon footprint of systems is becoming a dominating factor for AI's overall environmental impact (Section~\ref{sec:ai-carbon-footprint})~\cite{Gupta:HPCA:2021}.

\textbf{The Elephant in the Room.} 
Despite the positive societal benefits~\cite{ai-social-good}, the endless pursuit of achieving higher model quality has led to the exponential scaling of AI with significant energy and environmental footprint implications.
Although recent work shows the carbon footprint of training one large ML model, such as \emph{Meena}~\cite{Patterson:arxiv:2021}, is equivalent to 242,231 miles driven by an average passenger vehicle~\cite{GHG-calculator}, this is only one aspect; to fully understand the real environmental impact
we must consider the AI ecosystem \textit{holistically} going forward --- beyond looking at model training alone and by accounting for both \textit{operational} and \textit{embodied carbon footprint} of AI.
We must look at the ML pipeline end-to-end: data collection, model exploration and experimentation, model training, model optimization and run-time inference. 
The \textit{frequency of training} and \textit{scale} of each stage of the ML development cycle matter. 
From the systems perspective, the life cycle of ML software and system hardware, including manufacturing and operational use, must also be considered. 

\begin{figure*}[t]
    \centering
    \includegraphics[width=1\linewidth]{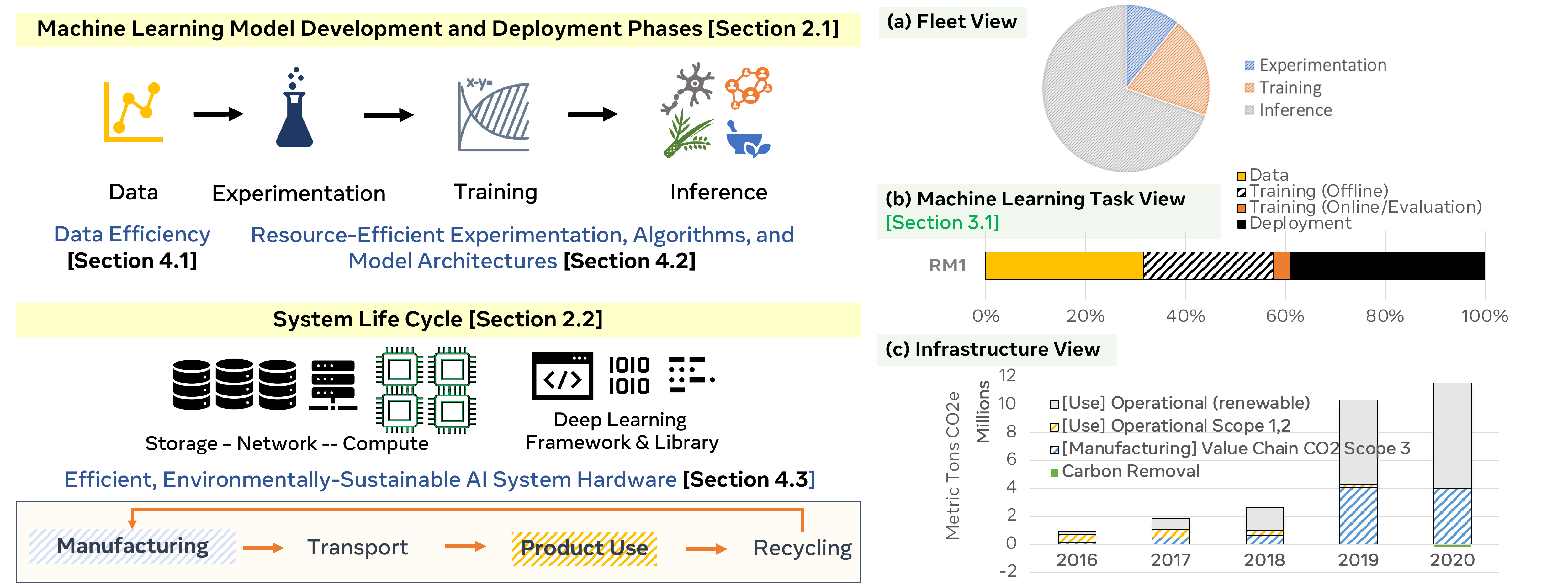}
    \caption{Model Development Phases over AI System Hardware Life Cycle: (a) At \fb, we observe a rough power capacity breakdown of \textbf{10:20:70} for AI infrastructures devoted to the three key phases --- \textbf{Experimentation}, \textbf{Training}, and \textbf{Inference}; (b) Considering the primary stages of the ML pipeline end-to-end, the energy footprint of RM1 is roughly \textbf{31:29:40} over \textbf{Data},  \textbf{Experimentation/Training}, and \textbf{Inference}; 
    (c) Despite the investment to neutralize the operational footprint with carbon-free energy, the overall data center electricity use continues to grow, demanding over 7.17 million MWh in 2020~\cite{facebook-sustainability-report}.}
    \label{fig:ml_lifecycle}
\end{figure*}

Optimizing across ML pipelines and systems life cycles end-to-end is a complex and challenging task. 
While training large, sparsely-activated neural networks improves model scalability, achieving higher accuracy at lower operational energy footprint~\cite{Patterson:arxiv:2021},
it can incur higher embodied carbon footprint from the increase in the system resource requirement.
Shifting model training and inference to data centers with carbon-free energy can reduce emissions; however, this approach may not scale to a broad set of use cases.
Infrastructure for carbon-free energy is limited by factors such as geography and available materials (e.g. rare metals), and takes significant economic resources and time to build.
In addition, as on-device learning becomes more ubiquitously adopted to improve data privacy, we can see more computation being shifted away from data centers to the edge, where access to renewable energy is limited. 

\textbf{A Holistic Approach.} This paper is the first to take a holistic approach to characterize the environmental footprint of AI computing from \textit{experimentation} and \textit{training} to \textit{inference}. We characterize the carbon footprint of AI computing by examining the model development cycle across industry-scale machine learning use cases at Facebook (Section~\ref{sec:model-life-cycle-analysis}). This is illustrated by the more than 800$\times$ operational carbon footprint reduction achieved through judicious hardware-software co-design for a Transformer-based universal language model.
Taking a step further, we present an end-to-end analysis for both operational \textit{and} embodied carbon footprint for AI training and inference (Section~\ref{sec:ai-carbon-footprint}).
Based on the industry experience and lessons learned, we chart out opportunities and important development directions across the dimensions of AI including --- data, algorithm, systems, metrics, standards, and best practices (Section~\ref{sec:optimization-opportunities}). 
We hope the key messages (Section~\ref{sec:takeaways}) and the insights in this paper can inspire the community to advance the field of AI in an environmentally-responsible manner.

%% file: tex/model-life-cycle.tex
\section{Model Development Phases and AI System Hardware Life Cycle}
\label{sec:model-life-cycle-analysis}

Figure~\ref{fig:ml_lifecycle} depicts the major development phases for ML --- \textbf{Data Processing}, \textbf{Experimentation}, \textbf{Training}, and \textbf{Inference} (Section~\ref{sec:ml-model-lifecycle}) --- over the life cycle of AI system hardware (Section~\ref{sec:ml-hardware-lifecycle}).
Driven by distinct objectives of AI research and advanced product development, infrastructure is designed and built specifically to maximize data storage and ingestion efficiency for the phase of \textbf{Data Processing}, developer efficiency for the phase of \textbf{Experimentation}, training throughput efficiency for the phase of \textbf{Training}, and tail-latency bounded throughput efficiency for \textbf{Inference}.

\subsection{Machine Learning Model Development Cycle}
\label{sec:ml-model-lifecycle}

ML researchers extract features from data during the \textbf {Data Processing} phase and apply weights to individual features based on feature importance to the model optimization objective.
During \textbf{Experimentation}, the researchers design, implement and evaluate the quality of proposed algorithms, model architectures, modeling techniques, and/or training methods for determining model parameters.
This model exploration process is computationally-intensive. A large collection of diverse ML ideas are explored simultaneously at-scale. 
Thus, during this phase, we observe unique system resource requirements from the large pool of training experiments. 
Within \fb's ML research cluster, 50\% (p50) of ML training experiments take up to 1.5 GPU days while 99\% (p99) of the experiments complete within 24 GPU days. There are a number of large-scale, trillion parameter models which require over 500 GPUs days.

Once a ML solution is determined as promising, it moves into \textbf{Training} where the ML solution is evaluated using extensive production data --- data that is \textit{more recent}, is \textit{larger in quantity}, and contains \textit{richer features}.
The process often requires additional hyper-parameter tuning. 
Depending on the ML task requirement, the models can be trained/re-trained at different frequencies. For example, models supporting \fb's \textit{Search} service were trained at an hourly cadence whereas the \textit{Language Translation} models were trained weekly~\cite{Hazelwood:hpca:2018}.
A p50 production model training workflow takes 2.96 GPU days while a training workflow at p99 can take up to 125 GPU days.

Finally, for \textbf{Inference}, the best-performing model is deployed, producing trillions of daily predictions to serve billions of users worldwide.
The total compute cycles for inference predictions are expected to exceed the corresponding training cycles for the deployed model. 

\subsection{Machine Learning System Life Cycle}
\label{sec:ml-hardware-lifecycle}

Life Cycle Analysis (LCA) is a common methodology to assess the carbon emissions over the product life cycle. There are four major phases: \textit{manufacturing}, \textit{transport}, \textit{product use}, and \textit{recycling}\footnote{Recycling is an important domain, for which the industry is developing a circular economy model to up-cycle system components --- design with recycling in mind.}. From the perspective of AI's carbon footprint analysis, \textit{\underline{manufacturing}} and \textit{\underline{product use}} are the focus. Thus, in this work, we consider the overall carbon footprint of AI by including \textit{manufacturing} --- carbon emissions from building 
infrastructures specifically for AI (i.e., \textit{embodied carbon footprint}) and \textit{product use} --- carbon emissions from the use of AI (i.e., \textit{operational carbon footprint}).

While quantifying the exact breakdown between operational and embodied carbon footprint is a complex process, we estimate the significance of embodied carbon emissions using \fb’s Greenhouse Gas (GHG) emission statistics\footnote{Facebook  Sustainability Data: \url{https://sustainability.fb.com/report/2020-sustainability-report/}.}. 
\textit{In this case, more than 50\% of \fb’s emissions owe to its value chain --- Scope 3 of \fb's GHG emission}. As a result, a significant embodied carbon cost is paid upfront for every system component brought into \fb's fleet of datacenters, where AI is the biggest growth driver.

%% file: tex/cf-characterization.tex
\section{AI Computing's Carbon Footprint}
\label{sec:ai-carbon-footprint}

\begin{figure}[t]
    \centering
    \includegraphics[width=\linewidth]{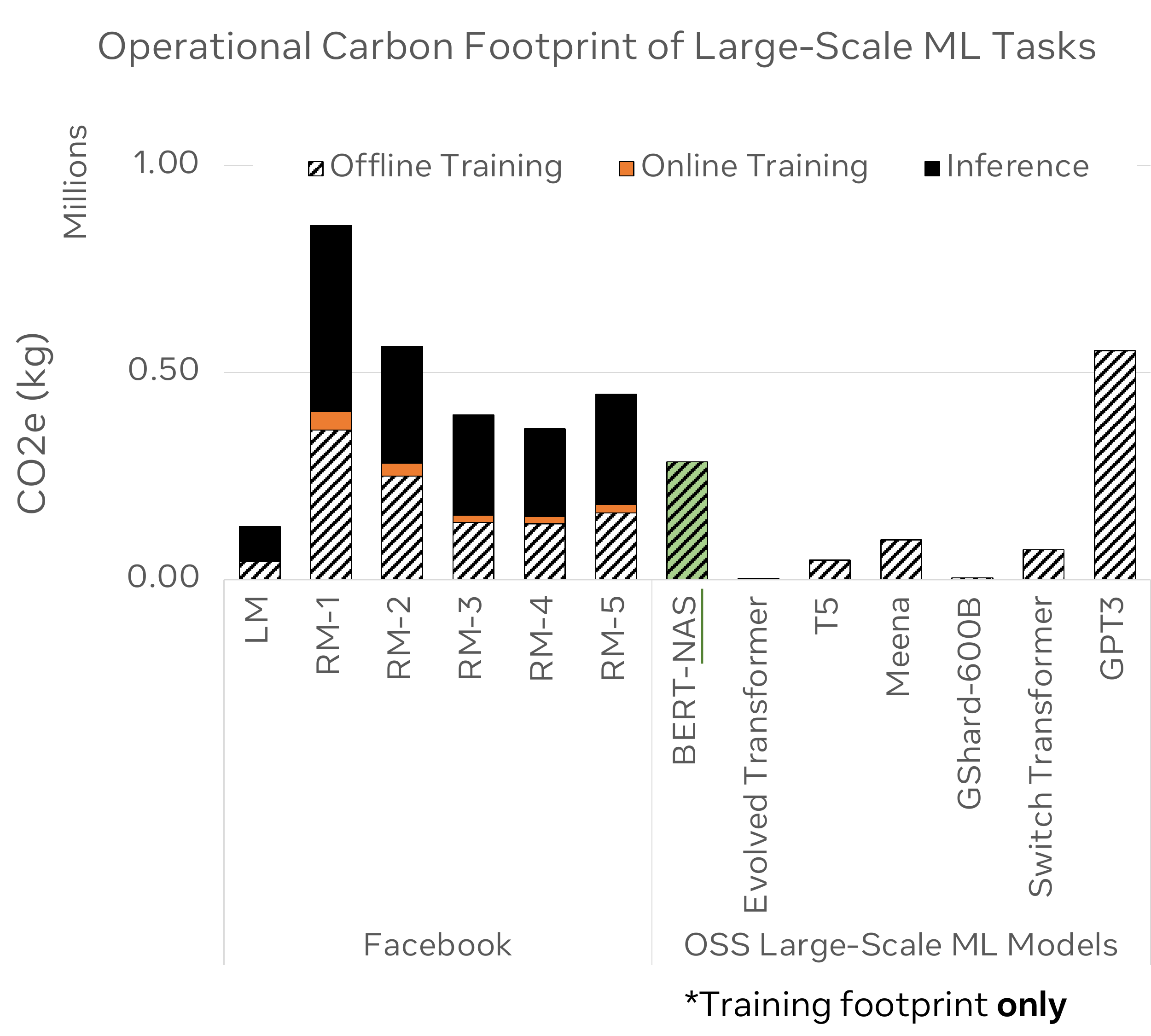}
    \caption{The carbon footprint of the LM model is dominated by Inference whereas, for RM1 -- RM5, the carbon footprint of Training versus Inference is roughly equal. The average carbon footprint for ML training tasks at Facebook is 1.8 times larger than that of Meena used in modern conversational agents and 0.3 times of GPT-3's carbon footprint. Carbon footprint for inference tasks is included for models that are used in production. Note: the operational carbon footprint of AI does not correlate with the number of model parameters. The OSS large-scale ML tasks are based on the vanilla model architectures from~\cite{Patterson:arxiv:2021} and may not be reflective of production use cases.}
    \label{figure:cf-characterization}
\end{figure}

\subsection{Carbon Footprint Analysis for Industry-Scale ML Training and Deployment}


Figure~\ref{figure:cf-characterization} illustrates the operational carbon emissions for model training and inference across the ML tasks. 
We analyze six representative machine learning models in production at \fb\footnote{In total, the six models account for a vast majority of compute resources for the overall inference predictions at \fb, serving billions of users world wide.}.
\textbf{LM} refers to \fb's Transformer-based Universal Language Model for text translation~\cite{XLM-r}.
\textbf{RM1} -- \textbf{RM5} represent five unique deep learning recommendation and ranking models for various Facebook  products~\cite{Naumov:arxiv:2019,Gupta:hpca:2020}. 

 We compare the carbon footprint of \fb's production ML models with seven large-scale, open-source (OSS) models: BERT-NAS, T5, Meena, GShard-600B, Switch Transformer, and GPT-3. 
 Note, we present the operational carbon footprint of the OSS model training from~\cite{Strubell:arxiv:2019,Patterson:arxiv:2021}. The operational carbon footprint results can vary based on the exact AI systems used and the carbon intensity of the energy mixture. Models with more parameters do not necessarily result in longer training time nor higher carbon emissions. Training the Switch Transformer model equipped with 1.5 trillion parameters~\cite{Fedus:switch-transformer:2021} produces significantly less carbon emission than that of GPT-3 (750 billion parameters)~\cite{brown:arxiv:2020}. This illustrates the carbon footprint advantage of operationally-efficient model architectures.
 
 \begin{figure}[t]
    \centering
    \includegraphics[width=\linewidth]{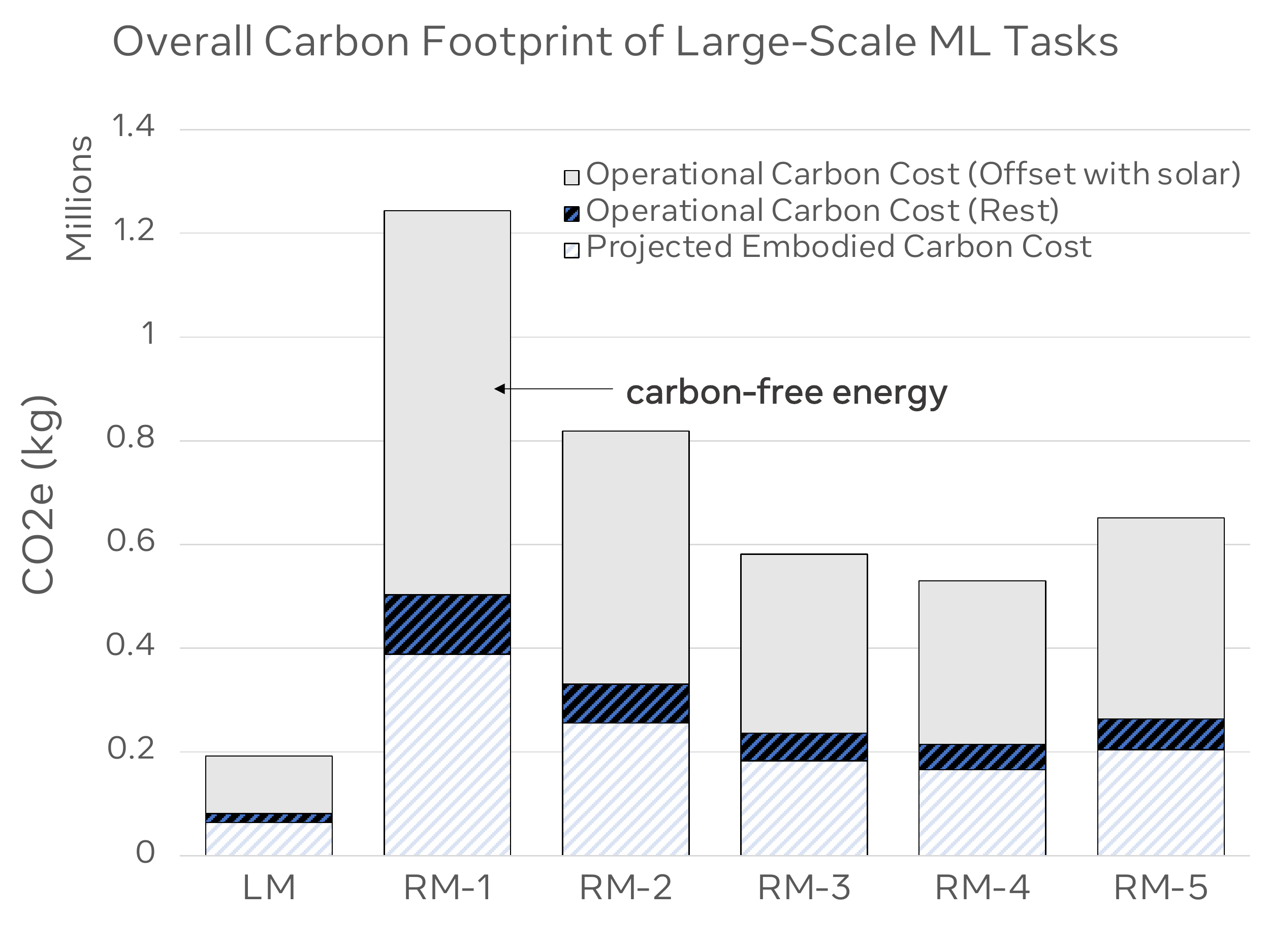}
    \caption{When considering the overall life cycle of ML models and systems in this analysis, manufacturing carbon cost is roughly 50\% of the (location-based) operational carbon footprint of large-scale ML tasks (Figure~\ref{figure:cf-characterization}). Taking into account carbon-free energy, such as solar, the operational energy consumption can be significantly reduced, leaving the manufacturing carbon cost as the dominating source of AI's carbon footprint.}
    \label{figure:ops-vs-embodied}
\end{figure}

\vspace{+0.2cm}
\noindent{\textit{Both \textbf{Training} and \textbf{Inference} can contribute significantly to the overall carbon footprint of machine learning tasks at \fb.
The exact breakdown between the two phases varies across ML use cases.}}
\vspace{+0.2cm}

The overall operational carbon footprint is categorized into \textit{offline training}, \textit{online training}, and \textit{inference}. 
Offline training encompasses both experimentation and training models with historical data.
Online training is particularly relevant to recommendation models where parameters are continuously updated based on recent data.
The inference footprint represents the emission from serving production traffic.
The online training and inference emissions are considered over the period of offline training.
 For recommendation use cases, we find the carbon footprint is split evenly between training and inference. On the other hand, the carbon footprint of LM is dominated by the inference phase, using much higher inference resources (65\%) as compared to training (35\%).

\vspace{+0.2cm}
\noindent{\textit{Both \textbf{operational} and \textbf{embodied carbon emissions} can contribute significantly to the overall footprint of ML tasks}.}
\vspace{+0.2cm}

\textbf{Operational Carbon Footprint:} 
Across the life cycle of the Facebook  models shown in Figure~\ref{figure:cf-characterization}, the average carbon footprint is 1.8$\times$ higher than that of the open-source Meena model~\cite{google-meena} and one-third of GPT-3's training footprint.
To quantify the emissions of \fb's models we measure the total energy consumed, assume location-based carbon intensities for energy mixes,\footnote{Renewable energy and sustainability programs of \fb~\cite{facebook-sustainability-report}.} and use a data center Power Usage Effectiveness (PUE) of 1.1. 
In addition to model-level and hardware-level optimizations, \fb's renewable energy procurement~\cite{facebook-sustainability-report} programs mitigates these emissions. 

\textbf{Embodied Carbon Footprint:} To quantify the embodied carbon footprint of AI hardware, we use LCA (Section~\ref{sec:ml-hardware-lifecycle}). 
We assume GPU-based AI training systems have similar embodied footprint
as the production footprint of Apple's 28-core CPU with dual AMD Radeon GPUs (2000kg CO$_2$)~\cite{appleMacProMax}.
For CPU-only systems, we assume half the embodied emissions.
Based on the characterization of model training and inference at \fb, we assume an average utilization of 30-60\% over the 3- to 5-year lifetime for servers.
Figure~\ref{figure:ops-vs-embodied} presents the overall carbon footprint for the large scale ML tasks at \fb, spanning both operational and embodied carbon footprint. Based on the assumptions of location-based renewable energy availability, the split between the embodied and (location-based) operational carbon footprint is roughly 30\% / 70\% for the large scale ML tasks. Taking into account carbon-free energy, such as solar, the operational carbon footprint can be significantly reduced, leaving the manufacturing carbon cost as the dominating source of AI's carbon footprint.



\begin{figure}[t]
    \centering
    \includegraphics[width=1\linewidth]{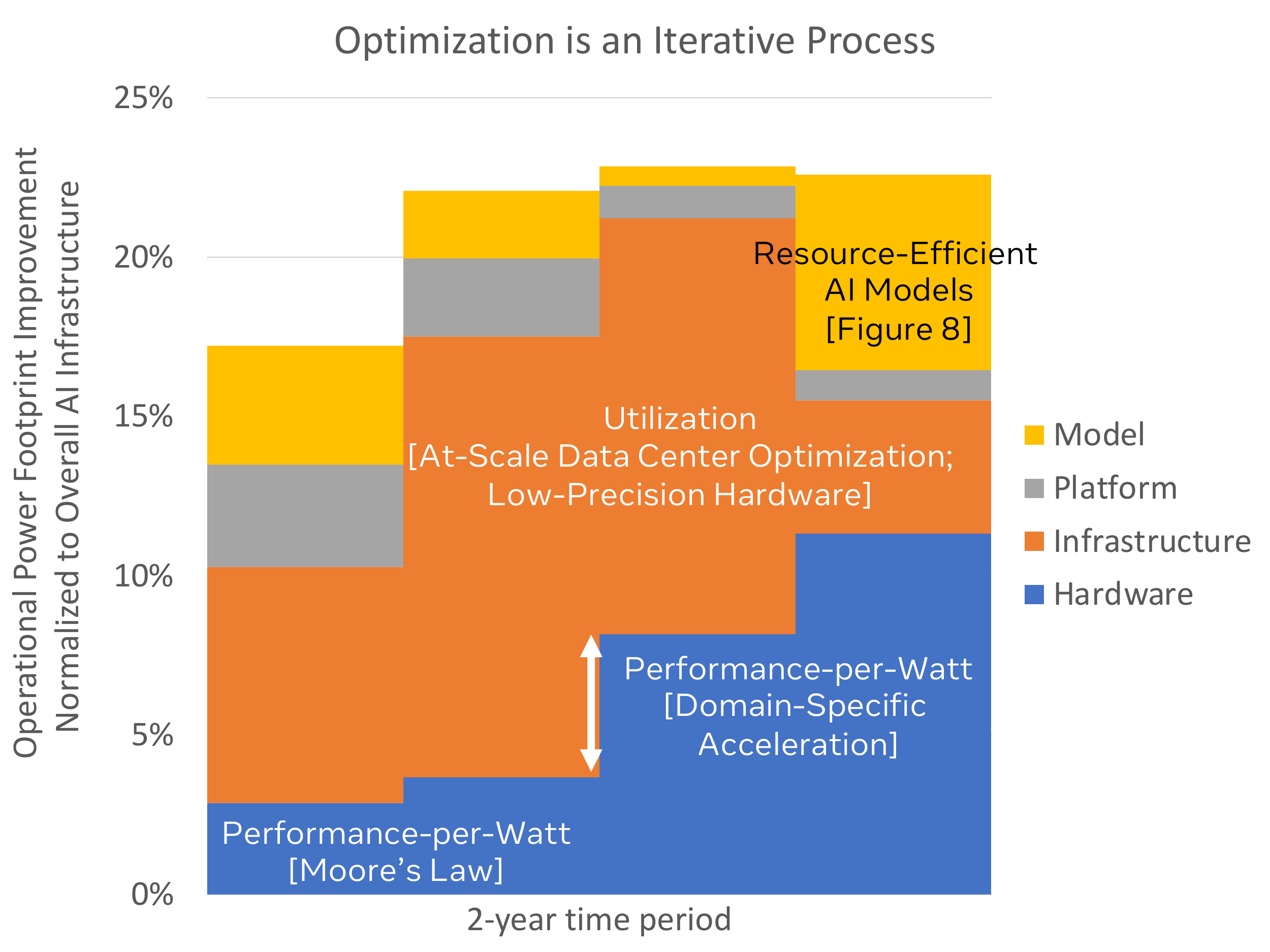}
    \caption{Optimization is an iterative process — we have achieved an average of 20\% operational energy footprint reduction every 6 months across the machine learning hardware-software stack.}  
    \label{fig:hw-sw-optimization}
\end{figure}

\subsection{Carbon Footprint Optimization from Hardware-Software Co-Design}
\label{sec:hw-sw-optimization}



\begin{figure}[t]
    \centering
    \includegraphics[width=\linewidth]{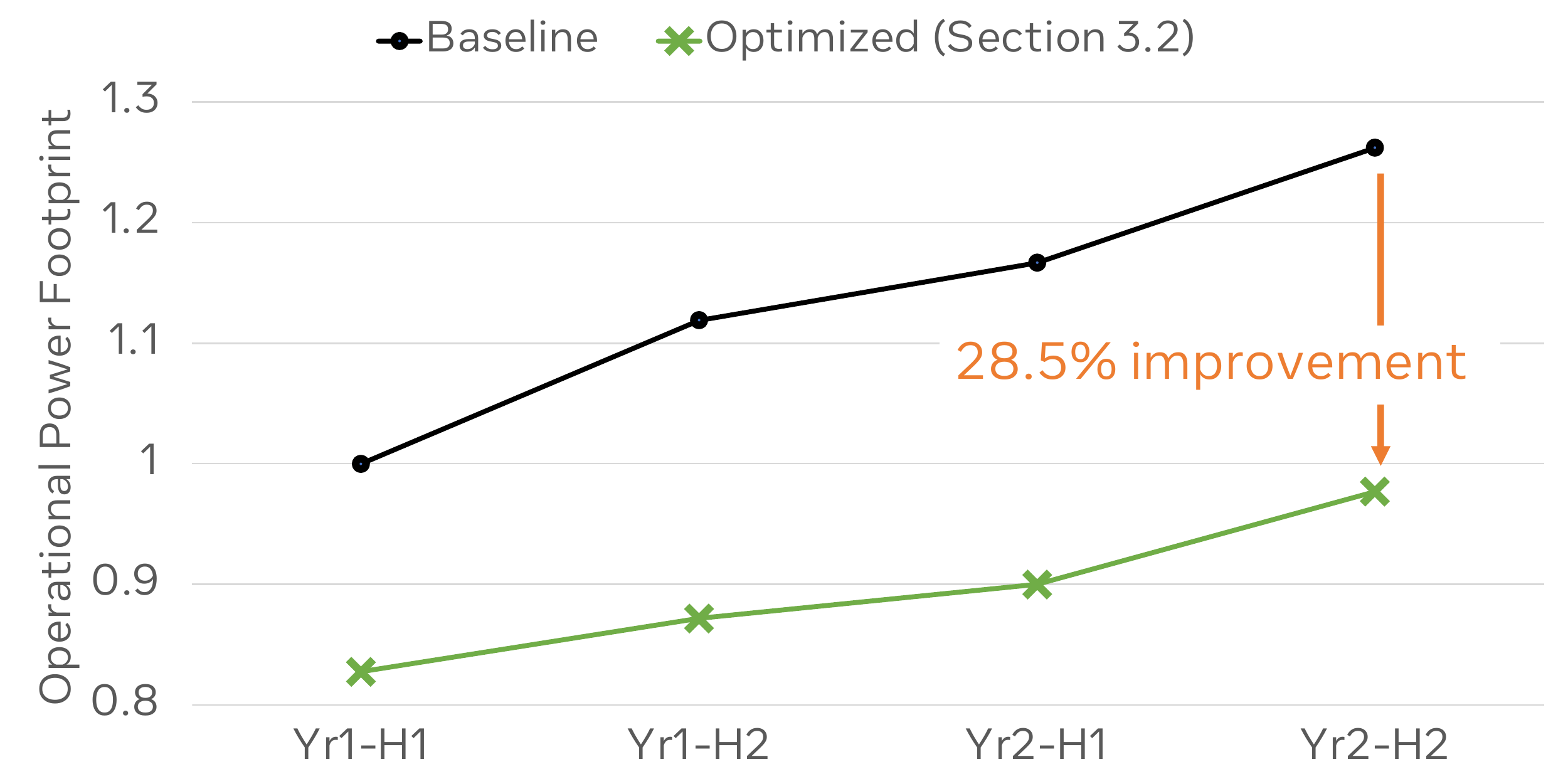}
    \caption{The iterative optimization process has led to 28.5\% operational energy footprint reduction over the two-year time period (Section~\ref{sec:hw-sw-optimization}). Despite the significant operational power footprint reduction, we continue to see the overall electricity demand for AI to increase over time --- an example of \textit{Jevon's Paradox}, where efficiency improvement stimulates additional novel AI use cases.}
    \label{fig:jevon-paradox}
\end{figure}

\vspace{+0.2cm}
\noindent{\textit{Optimization is an iterative process --- we reduce the power footprint across the machine learning hardware-software stack by ~20\% every 6 months. But at the same time, AI infrastructure continued to scale out. The net effect, with Jevon's Paradox, is a 28.5\% operational power footprint reduction over two years (Figure~\ref{fig:jevon-paradox}).}}
\vspace{+0.2cm}

\textbf{Optimization across AI Model Development and System Stack over Time:}
Figure~\ref{fig:hw-sw-optimization} shows the operational power footprint reduction across \fb's AI fleet over two years. 
The improvement come from four areas of optimizations: \textit{model} (e.g., designing resource-efficient models), \textit{platform} (e.g., PyTorch's support for quantization), \textit{infrastructure} (e.g., data center optimization and low-precision hardware), and \textit{hardware} (e.g., domain-specific acceleration).
Each bar illustrates the operational power reduction across \fb's AI fleet over 6-month period from each of the optimization areas.
The optimizations in aggregate provide, on average, a 20\% reduction in operational power consumption every six months. 
The compounded benefits highlight the need for cross-stack optimizations. 


\begin{figure}[t]
    \centering
    \includegraphics[width=\linewidth]{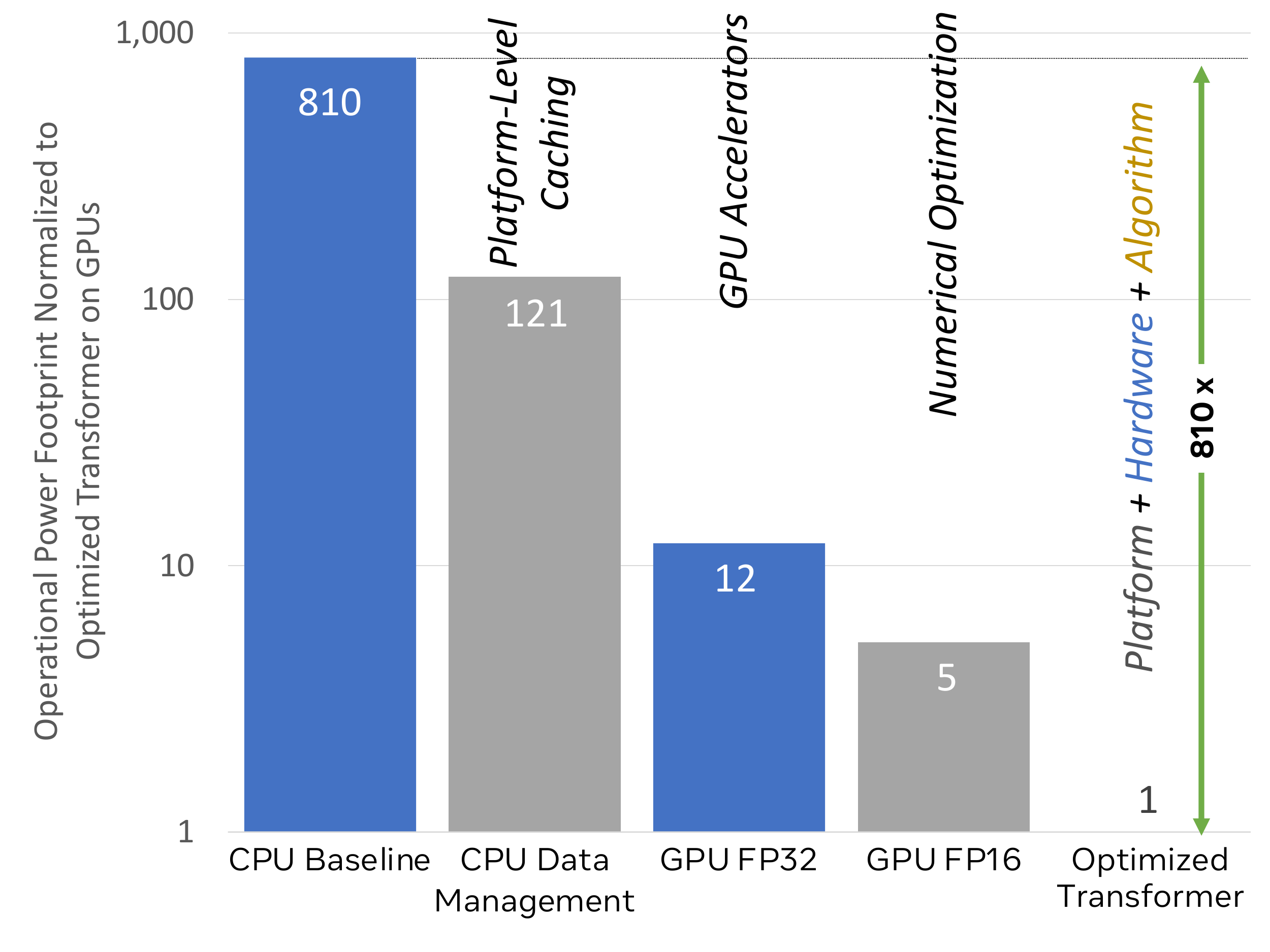}
    \caption{For the cross-lingual ML task (LM), the operational energy footprint can be significantly reduced by more than $800\times$ using \textit{platform-level caching}, \textit{GPUs}, \textit{low precision data format}, and \textit{additional algorithmic optimization}.}
    \label{fig:textray-optimization}
\end{figure}

\textbf{Optimizing the Carbon Footprint of \mbox{LMs}:} 
We dive into a specific machine learning task at \fb: language translation using a Transformer-based architecture (\mbox{LM}).
\mbox{LM} is designed based on the state-of-the-art cross-lingual understanding through self-supervision. 
Figure~\ref{fig:textray-optimization} analyzes the power footprint improvements over a collection of optimization steps for \mbox{LM}: \textit{platform-level caching}, \textit{GPU acceleration}, \textit{low precision format on accelerator}, and \textit{model optimization}. 
In aggregate the optimizations reduce the infrastructure resources required to serve \mbox{LM} at scale by over 800$\times$.
We outline the optimization benefits from each area below.
\begin{itemize}
    \item \textbf{Platform-Level Caching.} Starting with a CPU server baseline, application-level caching improves power efficiency by 6.7$\times$. These improvements are a result of pre-computing and caching frequently accessed embeddings for language translation tasks. Using DRAM and Flash storage devices as caches, these pre-computed embeddings can be shared across applications and use cases.
    \item \textbf{GPU acceleration.} In addition to caching, deploying LM across GPU-based specialized AI hardware unlocks an additional 10.1$\times$ energy efficiency improvement.

    \item \textbf{Algorithmic optimization.} Finally, algorithmic optimizations provide an additional 12$\times$ energy efficiency reduction. Halving precision (e.g., going from 32-bit to 16-bit operations) provides a 2.4$\times$ energy efficiency improvement on GPUs. Another 5$\times$ energy efficiency gain can be achieved by using custom operators to schedule encoding steps within a single kernel of the Transformer module, such as~\cite{faster-transformer}. 

\end{itemize}

\textbf{Optimizing the Carbon Footprint of \mbox{RMs}:}
The LM analysis is used as an example to highlight the optimization opportunities available with judicious cross-stack, hardware/software optimization. In addition to optimizing the carbon footprint for the language translation task,
we describe additional optimization techniques tailored for ranking and recommendation use cases. 


A major infrastructure challenge faced by deep learning RM training and deployment (\textbf{RM1} -- \textbf{RM5}) is the fast-rising memory capacity and bandwidth demands (Figure~\ref{fig:data-model-system-growth}). There are two primary sub-nets in a RM: the dense fully-connected (FC) network and the sparse embedding-based network. The FC network is constructed with multi-layer perceptions (MLPs), thus computationally-intensive. The embedding network is used to project hundreds of sparse, high-dimensional features to low-dimension vectors. It can easily contribute to over 95\% of the total model size. For a number of important recommendation and ranking use cases, the embedding operation dominates the inference execution time~\cite{Gupta:hpca:2020,Ke:isca:2020}.

To tackle the significant memory capacity and bandwidth requirement, we deploy model quantization for RMs~\cite{deng:ieee-micro-2021}. 
Quantization offers two primary efficiency benefits: the low-precision data representation reduces the amount of computation requirement and, at the same time, lowers the overall memory capacity need. 
By converting 32-bit floating-point numerical representation to 16-bit, we can reduce the overall RM2 model size by 15\%. This has led to 20.7\% reduction in memory bandwidth consumption.
Furthermore, the memory capacity reduction enabled by quantization unblocks novel systems with lower on-chip memory. For example, for RM1, quantization has enabled RM deployment on highly power-efficient systems with smaller on-chip memory, leading to an end-to-end inference latency improvement of 2.5 times.

\subsection{Machine Learning Infrastructures at Scale}

\textbf{ML Accelerators:}
GPUs are the de-facto training accelerators at \fb, contributing to significant power capacity investment in the context of \fb's fleet of datacenters. However, GPUs can be severely under-utilized during both the ML Experimentation and Training phases (Figure~\ref{fig:gpu-utilization})~\cite{Wesolowski:ieee-micro:2021}. To amortize the upfront embodied carbon cost of every accelerator deployed into \fb’s datacenters, maximizing accelerator utilization is a must. 

\textbf{Efficiency of Scale:} The higher throughput performance density achieved with ML accelerators reduces the total number of processors deployed into datacenter racks. This leads to more effective amortization of shared infrastructure overheads. Furthermore, datacenter capacity is not only limited by physical space but also power capacity --- higher operational power efficiency directly reduces the inherited carbon cost from manufacturing of IT infrastructures and datacenter buildings. 


\textbf{At-Scale Efficiency Optimization for Facebook Data Centers:}
Servers in Facebook data center fleets are customized for internal workloads only --- machine learning tasks~\cite{Hazelwood:hpca:2018} or not~\cite{Sriraman:isca:2019,Sriraman:asplos:2020}. Compared to public cloud providers, this puts Facebook at a unique position for at-scale resource management design and optimization. First, \fb customizes server SKUs --- compute, memcached, storage tiers and ML accelerators --- to maximize performance and power efficiency. Achieving a Power Usage Effectiveness (PUE) of about 1.10, \fb's data centers are about 40\% more efficient than small-scale, typical data centers. 

Furthermore, the large-scale deployment of servers of different types provides an opportunity to build performance measurement and optimization tools to ensure high utilization of the underlying infrastructure. For data center fleets in different geographical regions where the actual server utilization exhibits a diurnal pattern, Auto-Scaling frees the over-provisioned capacity during off-peak hours, by up to 25\% of the web tier’s machines~\cite{Tang:osdi:2020}. By doing so, it provides opportunistic server capacity for others to use, including offline ML training. Furthermore, static power consumption plays a non-trivial role in the context of the overall data center electricity footprint. This motivates more effective processor idle state management.

\textbf{Carbon-Free Energy:} Finally, over the past years, \fb has invested in carbon free energy sources to neutralize its operational carbon footprint~\cite{facebook-sustainability-report}. 
Reaching net zero emissions entails matching every unit of energy consumed by data centers with 100\% renewable energy purchased by \fb. Remaining emissions are offset with various sustainability programs, further reducing the operational carbon footprint of AI computing at \fb. As Section~\ref{sec:systems} will later show, \textit{more can be done}.

\subsection{Going Beyond Efficiency Optimization}

Despite the opportunities for optimizing energy efficiency and reducing environmental footprint at scale, there are many reasons why we must care about scaling AI in a more environmentally-sustainable manner. AI growth is multiplicative beyond current industrial use cases. Although domain-specific architectures improve the operational energy footprint of AI model training by more than 90\%~\cite{Patterson:arxiv:2021}, these architectures require more system resources, leading to larger embodied carbon footprints.

\begin{figure}[t]
    \centering
    \includegraphics[width=\linewidth]{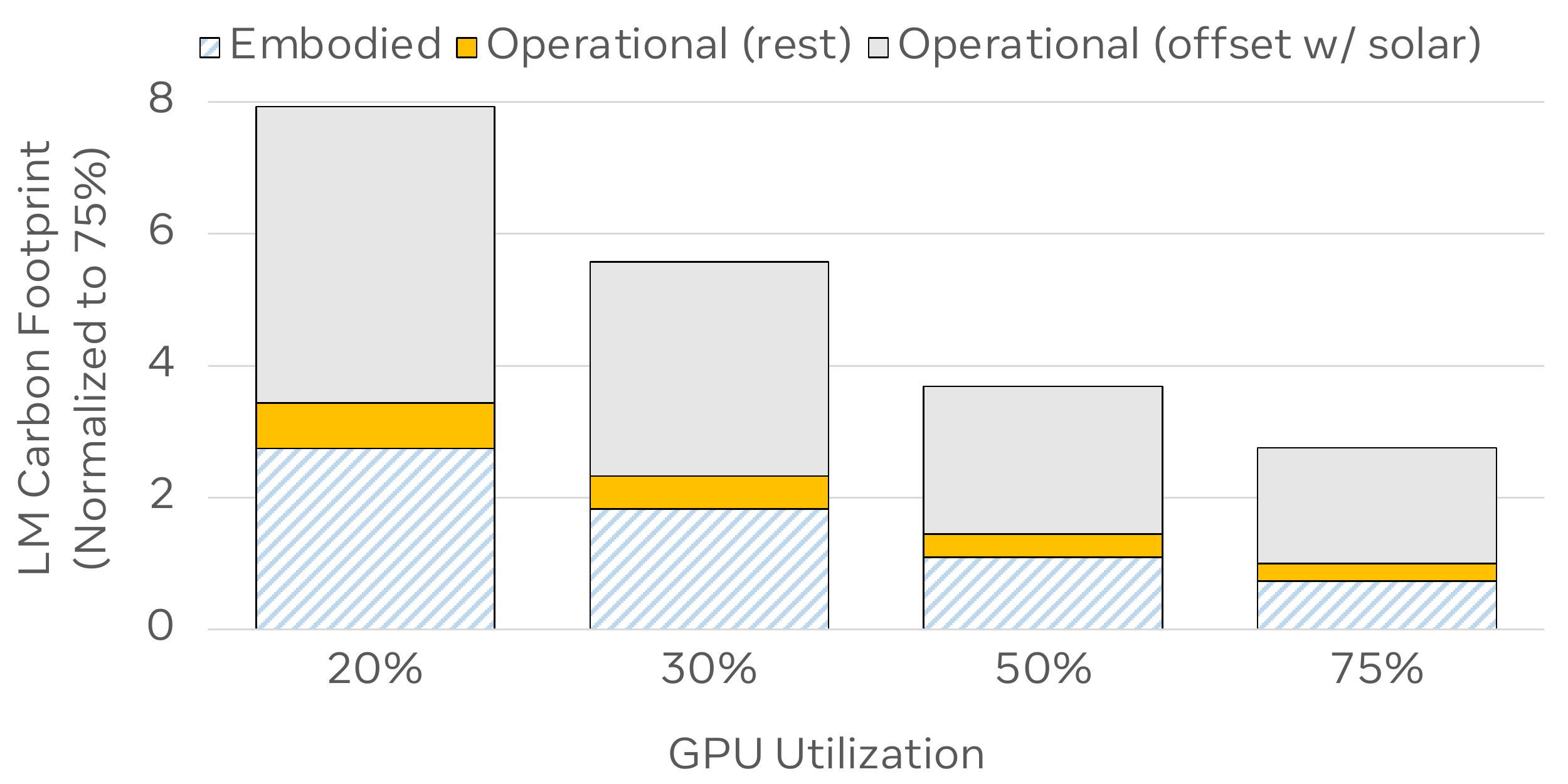}
    \vspace{-0.5cm}
    \caption{As accelerator utilization improves over time, both operational and embodied carbon footprints of AI improve. Carbon-free energy helps reduce the operational carbon footprint, making embodied carbon cost the dominating factor. 
    To reduce the rising carbon footprint of AI computing at-scale, we must complement efficiency and utilization optimization with novel approaches to reduce the remaining embodied carbon footprint of AI systems.}
    \label{figure:server-utilization}
    \vspace{-0.25cm}
\end{figure}
While shifting model training and inference to data centers with carbon-free energy sources can reduce emissions, the solution may not scale to all AI use cases.
Infrastructure for carbon free energy is limited by rare metals and materials, and takes significant economic resources and time to build. 
Furthermore, the carbon footprint of federated learning and optimization use cases at the edge is estimated to be similar to that of training a Transformer Big model (Figure~\ref{fig:fl_carbon}). As on-device learning becomes more ubiquitously adopted to improve data privacy, we expect to see more computation being shifted away from data centers to the edge, where access to renewable energy may be limited. The edge-cloud space for AI poses interesting design opportunities (Section~\ref{sec:systems}). 


\textit{The growth of AI in all dimensions outpaces the efficiency improvement at-scale.}
Figure~\ref{figure:server-utilization} illustrates that, as GPU utilization is improved (x-axis) for LM training on GPUs, both embodied and operational carbon emissions will reduce. Increasing GPU utilization up to 80\%, the overall carbon footprint decreases by 3$\times$. 
Powering AI services with renewable energy sources 
can further reduce the overall carbon footprint by a factor of 2. Embodied carbon cost becomes the dominating source of AI's overall carbon footprint. To curb the rising carbon footprint of AI computing at-scale (Figure~\ref{fig:jevon-paradox} and Figure~\ref{figure:server-utilization}), \textit{we must look beyond efficiency optimization and complement efficiency and utilization optimization with efforts to tackle the remaining embodied carbon footprint of AI systems.}


%% file: tex/futureofai.tex
\section{A Sustainability Mindset for AI}
\label{sec:optimization-opportunities}


To tackle the environmental implications of AI's exponential growth (Figure~\ref{fig:data-model-system-growth}), the first key step requires ML practitioners and researchers to develop and adopt an \textit{sustainability mindset}. The solution space is wide open---while there are significant efforts looking at \textit{AI system and infrastructure efficiency} optimization, the \textit{AI data, experimentation, and training algorithm efficiency} space (Sections~\ref{sec:data}  and~\ref{sec:experimentation-training}) beyond system design and optimization (Section~\ref{sec:systems}) is less well explored. 
We cannot optimize what cannot be measured --- telemetry to track the carbon footprint of AI technologies must be adopted by the community (Section~\ref{sec:metrics}). 
We synthesize a number of important directions to scale AI in a \textit{sustainable} manner and to minimize the environmental impact of AI for the next decades.

The field of AI is currently primarily driven by research that seeks to maximize model accuracy --- \textit{progress} is often used synonymously with improved prediction quality. This endless pursuit of higher accuracy over the decade of AI research has significant implications in computational resource requirement and environmental footprint.
To develop AI technologies responsibly, \textit{we must achieve competitive model accuracy at a fixed or even reduced computational and environmental cost}. 
Despite the recent calls-to-action~\cite{Strubell:arxiv:2019,Lacoste:arxiv:2019,Henderson:arxiv:2020,Bender:facct:2021,Patterson:arxiv:2021}, the overall community remains under-invested in research that aims at deeply understanding and minimizing the cost of AI. We conjecture the factors that may have contributed to the current state in Appendix~\ref{sec:appendix-efficiiency-mindset}.
To bend the exponential growth curve of AI and its environmental footprint, we must build a future where efficiency is an evaluation criterion for publishing ML research on computationally-intensive models beyond accuracy-related measures.

\subsection{Data Utilization Efficiency}
\label{sec:data}

\textbf{Data Scaling and Sampling:} \textit{No data is like more data} --- data scaling is the de-facto approach to increase model quality, where the primary factor for accuracy improvement is driven by the size and quality of training data, instead of algorithmic optimization. However, data scaling has significant environmental footprint implications. 
To keep the model training time manageable, overall system resources must be scaled with the increase in the data set size, resulting in larger embodied carbon footprint and operational carbon footprint from the data storage and ingestion pipeline and model training. 
Alternatively, if training system resources are kept fixed, data scaling increases training time, resulting in a larger operational energy footprint. 

When designed well, however, data scaling, sampling and selection strategies can improve the competitive analysis for ML algorithms, reducing the environmental footprint of the process (Appendix~\ref{sec:appendix-data-efficiency}). For instance, Sachdeva et al. demonstrated that intelligent data sampling with merely 10\% of data sub-samples can effectively preserve the relative ranking performance of different recommendation algorithms~\cite{Sachdeva:arxiv:2021}. This ranking performance is achieved with an average of 5.8 times execution time speedup, leading to significant operating carbon footprint reduction.

\textbf{Data Perishability:} Understanding key characteristics of data is fundamental to efficient data utilization for AI applications. \textit{Not all data is created equal} and data collected over time loses its predictive value gradually. 
Understanding the rate at which data loses its predictive value has strong implications on the resulting carbon footprint.
For example, natural language data sets can lose half of their predictive value in the time period of less than 7 years (the half-life time of data)~\cite{valavi:hbs:2020}. The exact half-life period is a function of context. If we were able to predict the half-life time of data, we can devise effective sampling strategies to subset data at different rates based on its half-life. 
By doing so, the resource requirement for the data storage and ingestion pipeline can be significantly reduced~\cite{Zhao:arxiv:2021} --- lower training time (operational carbon footprint) as well as storage needs (embodied carbon footprint).

\subsection{Experimentation and Training Efficiency}
\label{sec:experimentation-training}

The experimentation and training phases are closely coupled (Section~\ref{sec:model-life-cycle-analysis}).   
There is a natural trade-off between the investment in experimentation and the subsequent training cost (Section~\ref{sec:ai-carbon-footprint}).
\textbf{\emph{Neural architecture search} (NAS) and \emph{hyperparameter optimization} (HPO)} are techniques that automate the design space exploration. Despite their capability to discover higher-performing neural networks, NAS and HPO can be extremely resource-intensive, involving training many models, especially when using simple approaches. Strubell et al. show that grid-search NAS can incur over $3000\times$ environmental footprint overhead~\cite{Strubell:arxiv:2019}.
Utilizing much more sample-efficient NAS and HPO methods~\cite{Turner2021bbox,Ren2021NASsurvey} can translate directly into carbon footprint improvement.
In addition to reducing the number of training experiments, one can also reduce the training time of each experiment.
By detecting and \emph{stopping under-performing training workflows early}, unnecessary training cycles can be eliminated. 
    
\textbf{\emph{Multi-objective optimization}} explores the Pareto frontier of efficient model quality and system resource trade-offs. If used early in the model exploration process, it enables more informed decisions about \textit{which} model to train fully and deploy given certain infrastructure capacity. 
Beyond model accuracy and timing performance~\cite{Song:kdd:2020,Joglekar:kdd:2020,Tan:arxiv:2020,eriksson2021latencyNAS}, energy and carbon footprint can be directly incorporated into the cost function as optimization objectives 
to enable discovery of environmentally-friendly models. 
Furthermore, when training is decoupled from NAS, sub-networks tailoring to specialized system hardware can be selected \textit{without additional training}~\cite{cai:arxiv:2020,Stamoulis:arxiv:2019,Chen:arxiv:2021,Mellor:arxiv:2021}. Such approaches can significantly reduce the overall training time, however, at the expense of increased embodied carbon footprint.

Developing \textbf{\textit{resource-efficient model architectures}} fundamentally reduce the overall system capacity need of ML tasks.  
From the systems perspective, accelerator memory is scarce. 
However, DNNs, such as neural recommendation models, require significantly higher memory capacity and bandwidth~\cite{Acun:hpca:2021,Ke:isca:2020}. 
This motivates researchers to develop memory-efficient model architectures. For example, the Tensor-Train compression technique (TT-Rec) achieves more than 100$\times$ memory capacity reduction with negligible training time and accuracy trade-off~\cite{yin:mlsys:2021}. Similarly, the design space trade-off between memory capacity requirement, training time, and model accuracy is also explored in Deep Hash Embedding (DHE)~\cite{kang:kdd:2021}. While training time increases lead to higher operational carbon footprint, in the case of TT-Rec and DHE, the memory-efficient model architectures require significantly lower memory capacity while better utilizing the computational capability of training accelerators, resulting in lower embodied carbon footprint. 

Developing \textbf{\textit{efficient training algorithms}} is a long-time objective of research in optimization and numerical methods~\cite{nemirovskij1983problem}. 
Evaluations of optimization methods should account for \textit{all} experimentation efforts required to tune optimizer hyperparameters, not just the method performance after tuning~\cite{choi2019empirical,sivaprasad2020optimizer}. 
In addition, 
significant research has gone into algorithmic approaches to efficiently scale training~\cite{goyal2017accurate,ott2018scaling} by reducing communication cost via compression~\cite{alistarh2017qsgd,vogels2019powersgd}, pipelining~\cite{huang2019gpipe}, and sharding~\cite{rajbhandari2020zero,rasley2020deepspeed}. The advances have enabled efficient scaling to larger models and larger datasets. 
We expect efficient training methods to continue as an important domain.
While this paper has focused on supervised learning relying labeled data,   
algorithmic efficiency extends to other learning paradigms including self-supervised and semi-supervised learning (Appendix~\ref{sec:ssl}).


\subsection{Efficient, Environmentally-Sustainable AI Infrastructure and System Hardware}
\label{sec:systems}

To amortize the embodied carbon footprint, model developers and system architects must \textit{maximize the utilization of accelerator and system resources} when in use and \textit{prolong the lifetime of AI infrastructures}. 
Existing practices such as the move to domain-specific architectures at cloud scale~\cite{Jouppi:isca:2017,AWS-inferentia,Microsoft-graphcore} reduce AI computing’s footprint by consolidating computing resources at scale and by operating the shared infrastructures more environmentally-friendly with carbon free energy\footnote{We discuss additional important directions for building environmentally-sustainable systems in Appendix~\ref{sec:appendix-system-efficiency}, including datacenter infrastructure disaggregation; fault tolerant, resilient AI systems.}. 


\begin{figure}[t]
    \centering
    \includegraphics[width=\linewidth]{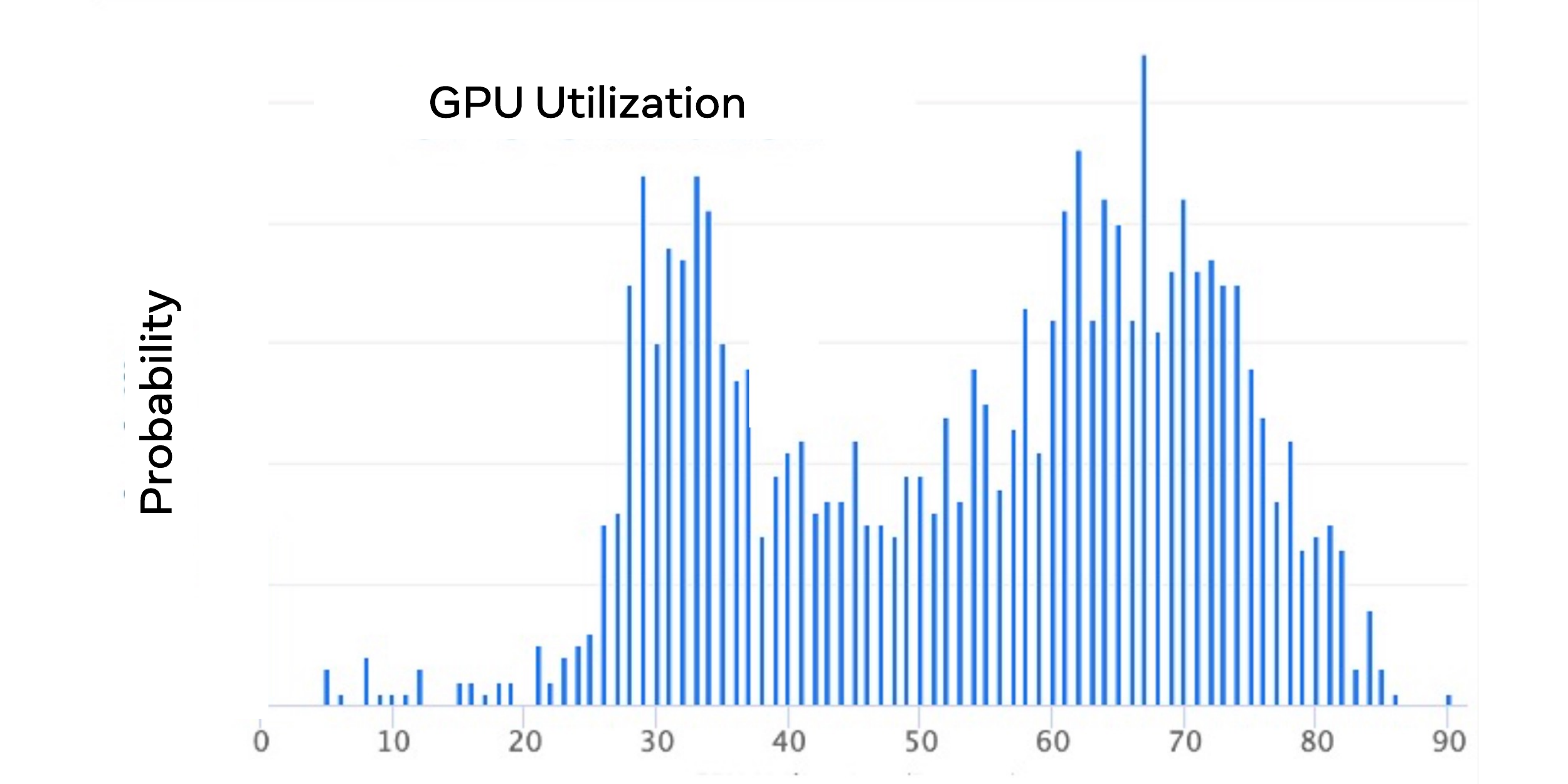}
    \caption{A vast majority of model experimentation (over tens of thousands of training workflows) utilizes GPUs at only 30-50\%, leaving room for utilization and efficiency improvements.}
    \label{fig:gpu-utilization}
\end{figure}

\textbf{Accelerator Virtualization and Multi-Tenancy Support:} Figure~\ref{fig:gpu-utilization} illustrates the utilization of GPU accelerators in \fb's research training infrastructure. A significant portion of machine learning model experimentation utilizes GPUs at only 30-50\%, leaving significant room for improvements to efficiency and overall utilization. Virtualization and workload consolidation technologies can help maximize accelerator utilization~\cite{GPU-vm}. Google's TPUs have also recently started supporting virtualization~\cite{TPU-vm}. Multi-tenancy for AI accelerators is gaining traction as an effective way to improve resource utilization, thereby amortizing the upfront embodied carbon footprint of customized system hardware for AI at the expense of potential operational carbon footprint increase~\cite{Gschwind:jrd:2017,Ghodrati:micro:2020,Kao:arxiv:2021,Jeon:usenix:2019,Yu:arxiv:2019}.

\textbf{Environmental Sustainability as a Key AI System Design Principle:}
Today, servers are designed to optimize performance and power efficiency. 
However, system design with a focus on operational energy efficiency optimization does not always produce the most environmentally-sustainable solution~\cite{jain:mobicom:2002,Chang:hotpower:2010,Gupta:HPCA:2021}.
With the rising embodied carbon cost and the exponential demand growth of AI, system designers and architects must re-think fundamental system hardware design principles to minimize computing’s footprint end-to-end, considering the entire hardware and ML model development life cycle. In addition to the respective performance, power, and cost profiles, the environmental footprint characteristics of processors over the generations of CMOS technologies, DDRx and HBM memory technologies, SSD/NAND-flash/HDD storage technologies can be orders-of-magnitude different~\cite{Bardon:iedm:2020}. Thus, designing AI systems with the least environmental impact requires explicit consideration of environmental footprint characteristics at the design time.

\textbf{The Implications of General-Purpose Processors, General-Purpose Accelerators, Reconfigurable Systems, and ASICs for AI:} 
There is a wide variety of system hardware choices for AI from general-purpose processors (CPUs), general-purpose accelerators (GPUs or TPUs), field-programmable gate arrays (FPGAs)~\cite{Putnam:ieee-micro-2015}, to application-specific integrated circuit (ASIC), such as Eyeriss~\cite{7551407}. 
The exact system deployment choice can be multifaceted --- 
the cadence of ML algorithm and model architecture evolution, the diversity of ML use cases and the respective system resource requirements, and the maturity of the software stack. 
While ML accelerator deployment brings a step-function improvement in \textit{operational energy efficiency}, it may not necessarily reduce the carbon footprint of AI computing overall. This is because of the upfront embodied carbon footprint associated with the different system hardware choices. 
From the environmental sustainability perspective, the optimal point depends on the compounding factor of operational efficiency improvement over generations of ML algorithms/models, deployment lifetime and embodied carbon footprint of the system hardware. Thus, to design for environmental sustainability, one must strike a careful balance between \textit{efficiency} and \textit{flexibility} and, at the same time, consider environmental impact as a key design dimension for next-generation AI systems.

\textbf{Carbon-Efficient Scheduling for AI Computing At-Scale:}
As the electricity consumption of hyperscale data centers continues to rise, data center operators have devoted significant investment to neutralize operational carbon footprint.
By operating large-scale computing infrastructures with carbon free energy, technology companies
are taking an important step to address the environmental implications of computing.
\textit{More can be done however}.

As the renewable energy proportion in the electricity grid increases, fluctuations in 
energy generation will increase due to the intermittent nature of renewable energy sources (i.e. wind, solar). Elastic carbon-aware workload scheduling techniques
can be used in and across datacenters to predict and exploit the intermittent energy generation patterns~\cite{radovanovic2021carbon}. However such scheduling algorithms might require server over-provisioning to allow for flexibility of shifting workloads to times when carbon-free energy is available.
Furthermore, any additional server capacity comes with manufacturing carbon cost
which needs to be incorporated into the design space. 
Alternatively, energy storage (e.g. batteries, pumped hydro, flywheels, molten salt) can be used to store renewable
energy during peak generation times for use during low generation times. 
There is an interesting design space to achieve 24/7 carbon-free AI computing.

\begin{figure}[t]
    \centering
    \includegraphics[width=\linewidth]{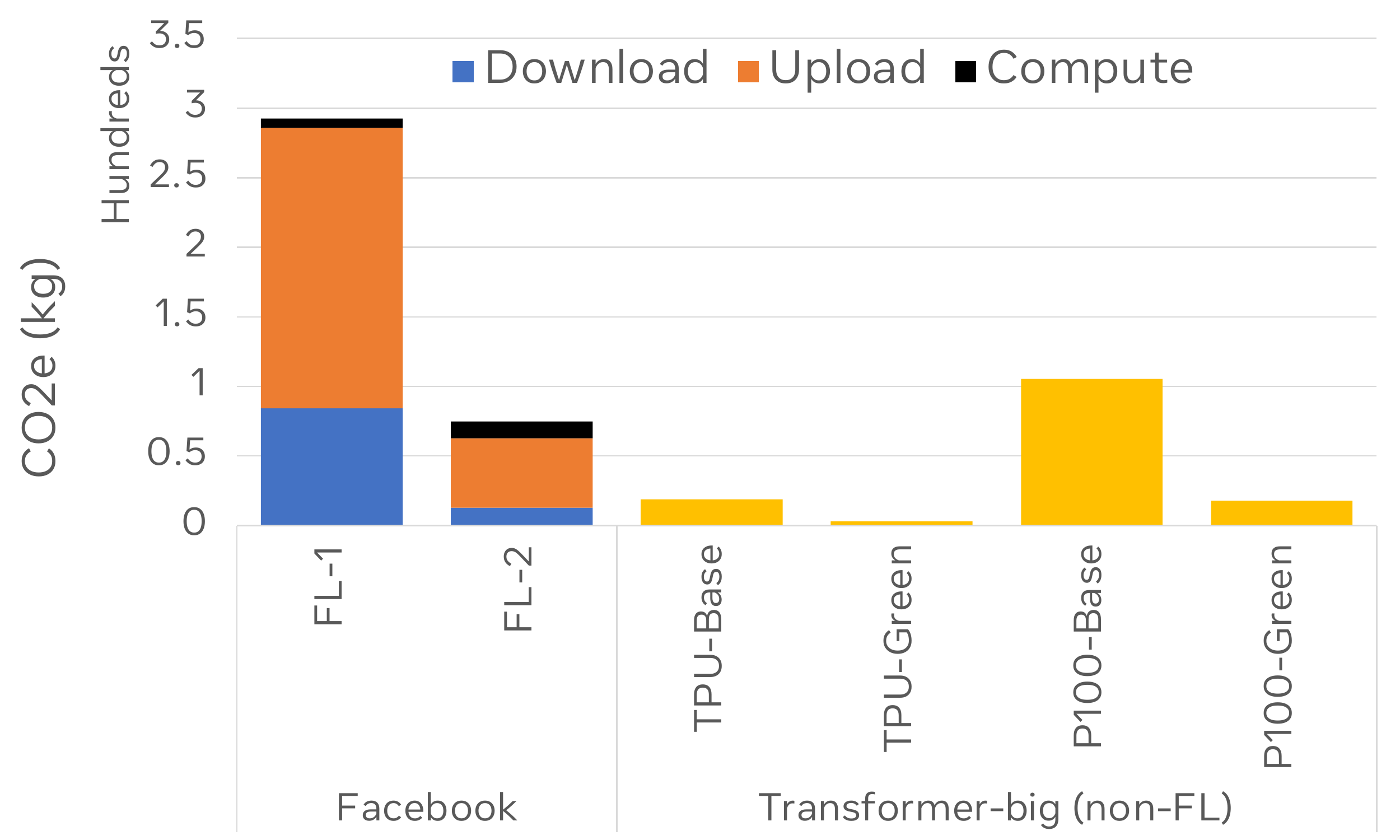}
    \caption{Federated learning and optimization can result in a non-negligible amount of carbon emissions, equivalent to the carbon footprint of training $Transformer_{Big}$~\cite{Patterson:arxiv:2021}. 
    FL-1 and FL-2 represent two production FL applications.
    P100-Base represents the carbon footprint of $Transformer_{Big}$ training on P100 GPU
    whereas TPU-base is $Transformer_{Big}$ training on TPU. P100-Green and TPU-Green consider renewable energy at the cloud (Methodology detail in Appendix~\ref{sec:appendix-system-efficiency}).}
    \label{fig:fl_carbon}
\end{figure}

\textbf{On-Device Learning}
On-device AI is becoming more ubiquitously adopted to enable model personalization~\cite{tinytl, fl_personalization,Bonawitz:arxiv:2019} while improving data privacy~\cite{gboard_prediction, gboard_ctr, gboard_emoji,huba2021papaya}, yet its impact in terms of carbon emission is often overlooked.
On-device learning emits non-negligible carbon. Figure~\ref{fig:fl_carbon} illustrates that the operational carbon footprint for training a small ML task using \emph{federated learning} (FL) is comparable to that of training an orders-of-magnitude larger Transformer-based model in a centralized setting.
As FL trains local models on client devices and periodically aggregates the model parameters for a global model, without collecting raw user data~\cite{gboard_prediction},
the FL process can emit non-negligible carbon at the edge due to both computation and wireless communication. 
%


It is important to reduce AI's environmental footprint at the edge. With the ever-increasing demand for on-device use cases over billions of client devices, such as teaching AI to understand the physical environment from the first-person perception~\cite{grauman:2021:ego4d} or personalizing AI tasks, the carbon footprint for on-device AI can add up to a dire amount quickly. Also, renewable energy is far more limited for client devices compared to datacenters.
Optimizing the overall energy efficiency of FL and on-device AI is an important first step~\cite{kim:micro:2021,kang:asplos:2017,kim:micro:2020,yang:arxiv:2017,Stamoulis:iccad:2018}. Reducing embodied carbon cost for edge devices is also important, as 
manufacturing carbon cost accounts for 74\% of the total footprint~\cite{Gupta:HPCA:2021} of client devices.
It is particularly challenging to amortize the embodied carbon footprint because client devices are often under-utilized~\cite{gao:ispass:2015}. 
%
%


\section{Call-to-Action}

\subsection{Development of Easy-to-Adopt Telemetry for Assessing AI's Environmental Footprint}
\label{sec:metrics}

While the open source community has started building tools to enable automatic measurement of AI training's environmental footprint~\cite{Lacoste:arxiv:2019,Henderson:arxiv:2020,codecarbon,Lottick:2019} and the ML research community requiring a broader impact statement for the submitted research manuscript, more can be done in order to incorporate efficiency and sustainability into the design process.
Enabling carbon accounting methodologies and telemetry that is easy to adopt is an important step to quantify the significance of our progress in developing AI technologies in an environmentally-responsible manner. While assessing the novelty and quality of ML solutions, it is crucial to consider sustainability metrics including \textit{energy consumption} and \textit{carbon footprint} along with measures of \textit{model quality} and \textit{system performance}. 

\textbf{Metrics for AI Model and System Life Cycles:} Standard carbon footprint accounting methods for AI's overall carbon footprint are at a nascent stage. We need simple, easy-to-adopt metrics to make fair and useful comparisons between AI innovations. Many different aspects must be accounted for, including the life cycles of both AI models (\textit{Data}, \textit{Experimentation}, \textit{Training}, \textit{Deployment}) and system hardware (\textit{Manufacturing} and \textit{Use}) (Section~\ref{sec:model-life-cycle-analysis}). 

In addition to incorporating an efficiency measure as part of leader boards for various ML tasks, data~\cite{kiela2021dynabench}, models\footnote{Papers with code: \url{https://paperswithcode.com/sota/image-classification-on-imagenet}}, training algorithms~\cite{hernandez2020efficiency}, environmental impact must also be considered and adopted by AI system hardware developers.
For example, MLPerf~\cite{Mattson:ieee-micro:2020,Reddi:ieee-micro:2021,mlperf:mobile} is the industry standard for ML system performance comparison. 
The industry has witnessed significantly higher system performance speedup, outstripping what is enabled by Moore's Law~\cite{mlperf-training,mlperf-inference}. Moreover, 
an algorithm efficiency benchmark is under development\footnote{\url{https://github.com/mlcommons/algorithmic-efficiency/}}. 
 The MLPerf benchmark standards can advance the field of AI in an environmentally-competitive manner by enabling the measurement of energy and/or carbon footprint.

\textbf{Carbon Impact Statements and Model Cards:} 
We believe it is important for all published research papers to disclose the operational \textit{and} embodied carbon footprint of proposed design; we are only at the beginning of this journey\footnote{\url{https://2021.naacl.org/ethics/faq/\#-if-my-paper-reports-on-experiments-that-involve-lots-of-compute-timepower}}. Note, while embodied carbon footprints for AI hardware may not be readily available, describing hardware platforms, the number of machines, total runtime used to produce results presented in a research manuscript is an important first step.
In addition, new models must be associated with a model card that, among other aspects of data sets and models~\cite{Mitchell:fat:2019}, describes the model’s overall carbon footprint to train and conduct inference. 


%% file: tex/takeaways.tex
\section{Key Takeaways}
\label{sec:takeaways}


\textbf{The Growth of AI:} Deep learning has witnessed an exponential growth in training data, model parameters, and system resources over the recent years (Figure~\ref{fig:data-model-system-growth}).
The amount of data for AI has grown by $2.4\times$, leading to $3.2\times$ increase in the data ingestion bandwidth demand at \fb.
\fb's recommendation model sizes have increased by $20\times$ between 2019 and 2021.
The explosive growth in AI use cases has driven $2.9\times$ and $2.5\times$ capacity increases for AI training and inference at Facebook over the recent 18 months, respectively.
The environmental footprint of AI is staggering (Figure~\ref{figure:cf-characterization}, Figure~\ref{figure:ops-vs-embodied}).

\textbf{A Holistic Approach:} To ensure an environmentally-sustainable growth of AI, we must consider the AI ecosystem holistically going forward.
We must look at the machine learning pipelines end-to-end --- data collection, model exploration and experimentation, model training, optimization and run-time inference (Section~\ref{sec:model-life-cycle-analysis}). 
The frequency of training and scale of each stage of the ML pipeline must be considered to understand salient bottlenecks to sustainable AI.
From the system's perspective, the life cycle of model development and system hardware, including \textit{manufacturing} and \textit{operational use}, must also be accounted for. 

\textbf{Efficiency Optimization:} 
Optimization across the axes of algorithms, platforms, infrastructures, hardware can significantly reduce the operational carbon footprint for the Transformer-based universal translation model by $810\times$. Along with other efficiency optimization at-scale, this has translated into 25.8\% operational energy footprint reduction over the two-year period. 
\textit{More must be done to bend the environmental impact from the exponential growth of AI} (Figure~\ref{fig:jevon-paradox} and Figure~\ref{figure:server-utilization}).

\textbf{An Sustainability Mindset for AI:} Optimization beyond efficiency across the software and hardware stack at scale is crucial to enabling future sustainable AI systems.
To develop AI technologies responsibly, we must achieve competitive model accuracy at a fixed or even reduced computational and environmental cost. We chart out potentially high-impact research and development directions across the \textit{data}, \textit{algorithms and model}, \textit{experimentation} and \textit{system hardware}, and \textit{telemetry} dimensions for AI at datacenters and at the edge (Section~\ref{sec:optimization-opportunities}). 

We must take a deliberate approach when developing AI research and technologies, considering the environmental impact of innovations and taking a responsible approach to technology development~\cite{wu:arxiv:2021}. That is, we need AI to be green and environmentally-sustainable.


%% file: tex/conclusion.tex
\section{Conclusion}
\label{sec:conclusion}

This paper is the first effort to explore the environmental impact of the super-linear trends for AI growth from a holistic perspective, spanning \textit{data}, \textit{algorithms}, and \textit{system hardware}. We characterize the carbon footprint of AI computing by examining the model development cycle across industry-scale ML use cases at Facebook and, at the same time, considering the life cycle of system hardware. Furthermore, we capture the operational and manufacturing carbon footprint of AI computing and present an end-to-end analysis for \textit{what} and \textit{how} hardware-software design and at-scale optimization can help reduce the overall carbon footprint of AI. We share the key challenges and chart out important directions across all dimensions of AI---data, algorithms, systems, metrics, standards, and best experimentation practices.
Advancing the field of machine intelligence must not in turn make climate change worse. We must develop AI technologies with a deeper understanding of the societal and environmental implications.


%% file: tex/ssl.tex
\emph{Self-supervised learning} (SSL) have received much attention in the research community in recent years. SSL methods train deep neural networks without using explicit supervision in the form of human-annotated labels for each training sample. Having humans annotate data is a time-consuming, expensive, and typically noisy process. SSL methods are typically used to train \emph{foundation models} --- models that can readily be fine-tuned using a small amount of labeled data on a down-stream task~\cite{bommasani2021opportunities}. SSL methods have been extremely successful for pre-training large language models, becoming the de-facto standard, and they have also attracted great interest in computer vision.

When comparing supervised and self-supervised methods, there is a glaring trade-off between having labels and the amount of computational overhead involved in pre-training. For example, Chen et al. report achieving 69.3\% top-1 validation accuracy with a ResNet-50 model after SSL pre-training for 1000 epochs on the ImageNet dataset and using the linear evaluation protocol, freezing the pre-trained feature extractor, and fine-tuning a linear classifier on top for 60 epochs using the full ImageNet dataset with all labels~\cite{chen2020simple}. In contrast, the same model typically achieves at least 76.1\% top-1 accuracy after 90 epochs of fully-supervised training. Thus, in this example, using labels and supervised training is worth a roughly 10$\times$ reduction in training effort, measured in terms of number of passes over the dataset.

Recent work suggests that incorporating even a small amount of labeled data can significantly bridge this gap. Assran et al. describe an approach called \emph{Predicting view Assignments With Support samples} (PAWS) for semi-supervised pre-training inspired by SSL~\cite{assran2021semi}. With access to labels for just 10\% of the training images in ImageNet, a ResNet-50 achieves 75.5\% top-1 accuracy after just 200 epochs of PAWS pre-training. Running on 64 V100 GPUs, this takes roughly 16 hours. Similar observations have recently been made for language model pre-training as well~\cite{dery2021should}.

Self-supervised pre-training potentially has advantages in that a single foundation model can be trained (expensive) but then fine-tuned (inexpensive), amortizing the up front cost across many tasks~\cite{bommasani2021opportunities}. Substantial additional research is needed to better understand the cost-benefit trade-offs for this paradigm.